\documentclass[twoside,11pt]{article}

\usepackage[preprint]{jmlr}

\jmlrheading{1}{2000}{1-48}{4/00}{10/00}{meila00a}{Hicham Janati, Marco Cuturi, Alexandre Gramfort}

\ShortHeadings{Spatio-temporal barycenters}{H. Janati, M. Cuturi, A. Gramfort}
\firstpageno{1}

\usepackage[english]{babel}
\usepackage[utf8]{inputenc} 

\usepackage{url}            
\usepackage{booktabs}       
\usepackage{amsfonts, mathrsfs, amssymb}       
\usepackage{nicefrac}       
\usepackage{microtype}      
\usepackage[export]{adjustbox}

\usepackage{subfigure,wrapfig}
\usepackage{dsfont}
\usepackage{amsmath}
\usepackage{stmaryrd}
\usepackage{xcolor}

\usepackage{breqn}

\usepackage{hyperref}
\usepackage{algorithm}
\usepackage{algorithmic}
\usepackage[labelfont=bf]{caption}
\usepackage{listings,newtxtt}

\usepackage{etoolbox}
\usepackage{subfiles}
\newtoggle{supplementary}
\toggletrue{supplementary}

\usepackage{nicefrac}       
\usepackage[labelfont=bf]{caption}
\usepackage{shortcuts}

\ifpdf
\hypersetup{
  pdftitle={Spatio-temporal barycenters},
  pdfauthor={H. Janati, M. Cuturi, and A. Gramfort}
}
\fi

\begin{document}

\title{Averaging Spatio-temporal Signals using\\ Optimal Transport and Soft Alignments}

\author{\name Hicham Janati \email hicham.janati@telecom-paris.fr \\
       \addr Télécom Paris,
       France
       \AND
       \name Marco Cuturi \email \email{marcocuturi@ensae.fr} \\
       \addr CREST ENSAE\\ Paris, France
       \AND \name Alexandre Gramfort \email
       alexandre.gramfort@inria.fr \\
       \addr Université Paris-Saclay, Inria, CEA, Palaiseau, France}

\editor{}

\maketitle

\begin{abstract}

    Several fields in science, from genomics to neuroimaging, require monitoring populations (measures) that evolve with time. These complex datasets, describing dynamics with both time and spatial components, pose new challenges for data analysis. We propose in this work a new framework to carry out \textit{averaging} of these datasets, with the goal of synthesizing a representative template trajectory from multiple trajectories. We show that this requires addressing three sources of invariance: shifts in time, space, and total population size (or mass/amplitude).
%
    Here we draw inspiration from dynamic time warping (DTW), optimal transport (OT) theory and its unbalanced extension (UOT) to propose a criterion that can address all three issues.
    This proposal leverages a smooth formulation of DTW (Soft-DTW) that is shown to capture temporal shifts, and UOT to handle both variations in space and size. Our proposed loss can be used to define spatio-temporal barycenters as Fréchet means. Using Fenchel duality, we show how these barycenters can be computed efficiently, in parallel, via a novel variant of entropy-regularized debiased UOT.  Experiments on handwritten letters and brain imaging data confirm our theoretical findings and illustrate the effectiveness of the proposed loss for spatio-temporal data.
\end{abstract}


\section{Introduction}
One of the most elementary operations in machine learning pipelines is to summarize data by aggregating several samples into one. On metric spaces, while the Euclidean mean is probably the most standard averaging tool, it is often not suited to applications where features are inherently structured independently of the data. This is for instance the case with spatio-temporal data where features correspond to fixed physical positions in space in some given moments in time. 
Spatio-temporal data can be seen as multivariate time-series where each (temporal) observation is defined in a space equipped with a physical notion of geometry. For instance, with videos displaying the motion of some object, the spatial features (pixels) are defined through a fixed rectangular grid while the temporal features are given in a chronological order. More generally, the spatial geometry can be defined on a predefined graph. This is the case with brain imaging data, where the measurements of neural activity are provided at each vertex of the triangulated mesh of the brain and across several time points~\citep{gramfort-etal:2011}. 

 Formally, each sample $\mu^i$ in our dataset can be represented as a time sequence of discrete non-negative measures $(\mu^i_t)_{t=1..T}$ defined in some finite set $\cX$. Our aim is to find a loss function $\cL$ through which the weighted Fréchet mean can be defined as:
\begin{equation}
\label{eq:frechet}
\bar{\mu} = \argmin_{\mu \in (\cM_+{(\cX)})^T} \sum_{i=1}^N w_i \cL(\mu^i, \mu)\enspace,
\end{equation}
where $(w_i){i=1..N}$ is a set of positive weights summing to one.

In the videos example, $\cX \eqdef \{x_1, \dots, x_p\}$ would be the set of pixels and $\mu^i_t \in \cM_+(\cX)$ would correspond to the $t$-th frame of the $i$-th video. Given that $\mu^i_t$ is discrete with a fixed support $\cX$, it can be identified with a non-negative vector of weights $\bp \in \bbR^{p}_+$ such that $\mu^i_t =  \sum_{k=1}^p \bp_k \delta_{x_k}$. Such a setting was considered in \citeauthor{janati20-aistats}'s proposal~\citeyear{janati20-aistats} where they defined $\cL$ in \eqref{eq:frechet} as an alignment soft-DTW metric operating on top of an OT cost. Intuitively, DTW is computed by finding the alignment between time points that has the lowest cost. If this cost is defined through OT, then the obtained optimal temporal alignment would pair time points that are \emph{spatially} similar to each other while capturing temporal differences.~\citet{janati20-aistats} propose to use Soft-DTW~\citep{cuturi17} -- rather than DTW~\citep{sakoe78} -- to benefit from additional properties notably a different behaviour with respect to time shifts~\citep{janati20-aistats}. In this paper, our work sees instead in Soft-DTW a differentiable loss that will be more adequate for the task of computing barycenters.

\paragraph{Related work}
On one hand, ignoring the temporal dimension, one could leverage the geometric properties of optimal transport (OT) metrics by computing OT barycenters (i.e using an OT loss for $\cL$) sequentially through time i.e averaging all the samples for each time point $t$ independently of the others. These geometrical differences between measures are captured through some pre-defined ground metric between the spatial features. Taking once again the videos example, this ground metric can be for instance given by the Euclidean distance between the coordinates of the pixels in their fixed rectangular frame. A straightforward use of OT with spatio-temporal data is to consider the transportation across space and time simultaneously via a customized ground metric. Motivated by OT between signed signals, this customization of the ground metric was previously studied by \cite{thorpe17} and introduced as TL$_p$ distances. However, this method ignores the chronology of the data and requires a difficult tuning of the tradeoff between spatial and temporal transport costs when designing the ground metric.  

Recently, \citet{vayer2020} proposed a new derivative of soft-DTW that learns the global invariances of the data that is very well adapted for data depicting trajectories that are invariant by rotations. This method however is not suited for comparing and averaging complex mass dynamics that cannot be modeled as a set of trajectories which is the case with meteorological or neuroimaging data for instance.
Inspired by the Gromov-Wasserstein framework, this proposal was later generalized by \citet{cohen2021aligning} to align and average time series in different spaces.

\paragraph{Challenges and contributions} 
Following \citet{janati20-aistats}, we propose to define the loss function $\cL$ using Soft-DTW and unbalanced optimal transport (UOT). This combination is straightforward since Soft-DTW requires a cost function of its own to align temporal observations across time series. Using UOT as acost function leads to the loss function coined STA: \emph{spatio-temporal alignement}. While evaluating STA can be easily carried out to perform clustering or metric based classification, computing its Fréchet mean comes with several challenges:

\begin{itemize}
    \item A classic bottleneck when using OT is scalability. 
    In ~\cite{janati20-aistats}, the differentiability of Soft-DTW is more or less hinted, but not experimented with. Attempting to solve \eqref{eq:frechet} naively via gradient descent turns out to be computationally infeasible as one gradient step requires computing $N T^2$ gradients of entropic OT, and thus running $N T^2$ Sinkhorn loops. Instead we propose to re-write the loss function using Fenchel duality so that one can update the spatio-temporal barycenter using the generalized Sinkhorn's algorithm for barycenters.
    \item As in \citep{janati20-aistats}, we use a debiased variant of unbalanced OT that is non-negative and suffers from less entropic blur. Starting from ~\citeauthor{janati20-icml}'s proposal~\citeyear{janati20-icml} to modify the Sinkhorn algorithm to compute \textit{debiased} OT barycenters (that cancel the blurring effect induced by entropic regularization), we extend their work and generalize it to the \textit{unbalanced} case, where the marginals have different sums.
    \item As is usual in mixed approaches that combine several metrics \citep{thorpe17, Damodaran2018DeepJDOTDJ}, setting hyperparameters that control the tradeoffs between these different contributions is challenging. We propose a heuristic to set the hyperparameters of OT and soft-DTW based on a maximum desired temporal shift.
    \item To find this heuristic, we derive tight inequalities that bound the growth of Delannoy numbers which can be of independent interest in Combinatorics.
\end{itemize}

\paragraph{Structure}
Section \ref{s:soft-dtw} provides background material on Soft-DTW, our contributions to Delannoy numbers that lead to a practical heuristic to set the Soft-DTW hyperparameter. Next in section \ref{s:ao}, we propose an alternating optimization algorithm to compute the Fréchet mean \eqref{eq:frechet}. This alternating algorithm reduces the problem to a sequence of temporally weighted Fréchet means of the Soft-DTW cost. In section \ref{s:ot}, we propose a generalized Sinkhorn algorithm to compute these Fréchet means for debiased unbalanced OT barycenters. Finally, we showcase the performance of STA averaging in Section \ref{s:experiments} in a forecasting experiment of handwritten characters and averaging of brain imaging data. 

\paragraph{Notation}
We denote by $\mathds 1_p$ the vector of ones in $\bbR^p$ and by $\intset{q}$ the set $\{1, \ldots, q\}$ for any integer $q \in \bbN$. The set of vectors in $\bbR^p$ with non-negative (resp. positive) entries is denoted by $ \bbR^p_+$ (resp. $\bbR^p_{++}$).  On matrices, $\log$, $\exp$ and the division operator are applied element-wise. We use $\odot$ for the element-wise multiplication between matrices or vectors. If $\bX$ is a matrix, $\bX_{i.}$ denotes its $i^{\text{th}}$ row and $\bX_{.j}$ its $j^{\text{th}}$ column. We define the Kullback-Leibler (KL) divergence between two positive vectors by $\kl(\bx, \by) = \langle \bx , \log(\bx / \by) \rangle + \langle \by - \bx, \mathds 1_p \rangle$ with the continuous extensions  $0\log(0 / 0) = 0 $ and $0 \log(0) = 0$. We also make the convention $\bx \neq 0 \Rightarrow \kl(\bx | 0) = +\infty$. The entropy of $\bx \in \bbR^p$ is defined as $H(\bx) = \langle \bx,\log(\bx) - \mathds 1_p \rangle $. The same definition applies for matrices with an element-wise double sum. The feasible set of binary matrices of $\bbR^{T_1 \times T_2}$ where only $\rightarrow, \downarrow, \searrow$ movements are allowed is denoted by $\cA_{T_1, T_2}$.


\section{Soft-DTW and Delannoy numbers}
\label{s:soft-dtw}
\subsection{Background}
\paragraph{Dynamic time warping}
Consider two multivariate time series $\bx \in \bbR^{p, T_1}$ and $\by \in \bbR^{p, T_2}$ with respective lengths $T_1, T_2$ and having observations in $\bbR^p$. DTW is defined through some pairwise distance matrix $\Delta(\bx, \by) \in \bbR^{T_1, T_2}$ between all their time points such that the cost of a given alignment function $\sigma: \intset{1, T_1} \to \intset{1, T_2}$ is equal to $\sum_{i=1}^{T_1} \Delta(\bx_i, \by_{\sigma(i)})$. To guarantee the preservation of the chronology of the data, $\sigma$ must be increasing and verify $\sigma(1) = 1$ and $\sigma(T_1) = T_2$. The resulting optimization problem is however better posed as a minimization of $\sum_{i=1}^{T_1}\sum_{j=1}^{T_2} A_{ij}\Delta(\bx_i, \by_j)$ over the set of binary alignments $A$ on the rectangular lattice $\intset{1, T_1} \times \intset{1, T_2}$ where no temporal back steps are allowed. This amounts to considering  binary matrices with a non-zero path linking the corners of the lattice $(1, 1)$ (upper left) and $(T_1, T_2)$ (bottom right) using $\rightarrow, \downarrow, \searrow $ steps exclusively~\citep{sakoe78}. Figure \ref{f:dtw-example} displays a toy example of such an alignment. Formally, DTW is defined as:
\begin{equation}
\label{eq:dtw}
\text{\textbf{dtw}}(\bx, \by; \Delta) = \min \{\langle \bA, \Delta(\bx, \by)\rangle, \bA \in \cA_{T_1, T_2}\} \enspace ,
\end{equation}
where $\langle., . \rangle$ denotes the Frobenius dot product.
 The binary nature of the constraint set in \eqref{eq:dtw} makes the DTW loss non-differentiable which is a major limitation when DTW is used as a loss function. To circumvent this issue, several authors introduced regularized variants of DTW \citep{saigo04, cuturi2007kernel, cuturi11, cuturi17}. Instead of selecting \emph{the} minimum cost alignment, Global Alignment Kernels (GAK) for instance \citep{saigo04, cuturi2007kernel,cuturi11} compute a weighted cost of all possible alignments with a certain smoothing hyperparameter. Similarly, the soft-minimum generalization approach of \cite{cuturi17} -- called soft-DTW -- provides a similar framework to that of GAK that includes DTW as a sub-case:
\begin{equation}
\label{eq:sdtw}
\sdtw(\bx, \by; \Delta) = {\softmin}_{\beta}\{\langle \bA, \Delta(\bx, \by)\rangle, \bA \in \cA_{T_1, T_2}\} \enspace,
\end{equation}
where the soft-minimum operator of a set $\cA$ with parameter $\beta \geq 0$ is defined as:
\begin{equation}
{\softmin}_{\beta}(\cA) =
\left\{\begin{array}{ll}
-\beta \log\left(\sum_{a \in \cA} e^{- \nicefrac{a}{\beta}}\right) \text{\quad if } \beta > 0\\
\min {\cA}  \text{\quad if } \beta = 0
\end{array} \right.
\end{equation}
In particular, softmin is continuous at 0 so that when $\beta \to 0$, $\sdtw$ approaches \textbf{dtw}.

\paragraph{Forward recursion}
Figure~\ref{f:dtw-example} illustrates two time series of images and their cost matrix $\Delta$. The path from (1, 1) to (5, 6) is an example of a feasible alignement in $\cA_{5, 6}$.
When $\beta = 0$, the soft-minimum is a minimum and $\sdtw$ falls back to the classical DTW metric. Nevertheless, it can still be computed using the dynamic program of Algorithm \ref{a:dynamicprogram} with a soft-min instead of min operator. 
\vspace{0.2cm}

\begin{minipage}{0.35\linewidth}
	\centering
	\includegraphics[width=\linewidth, trim={0 0 0 0},clip]{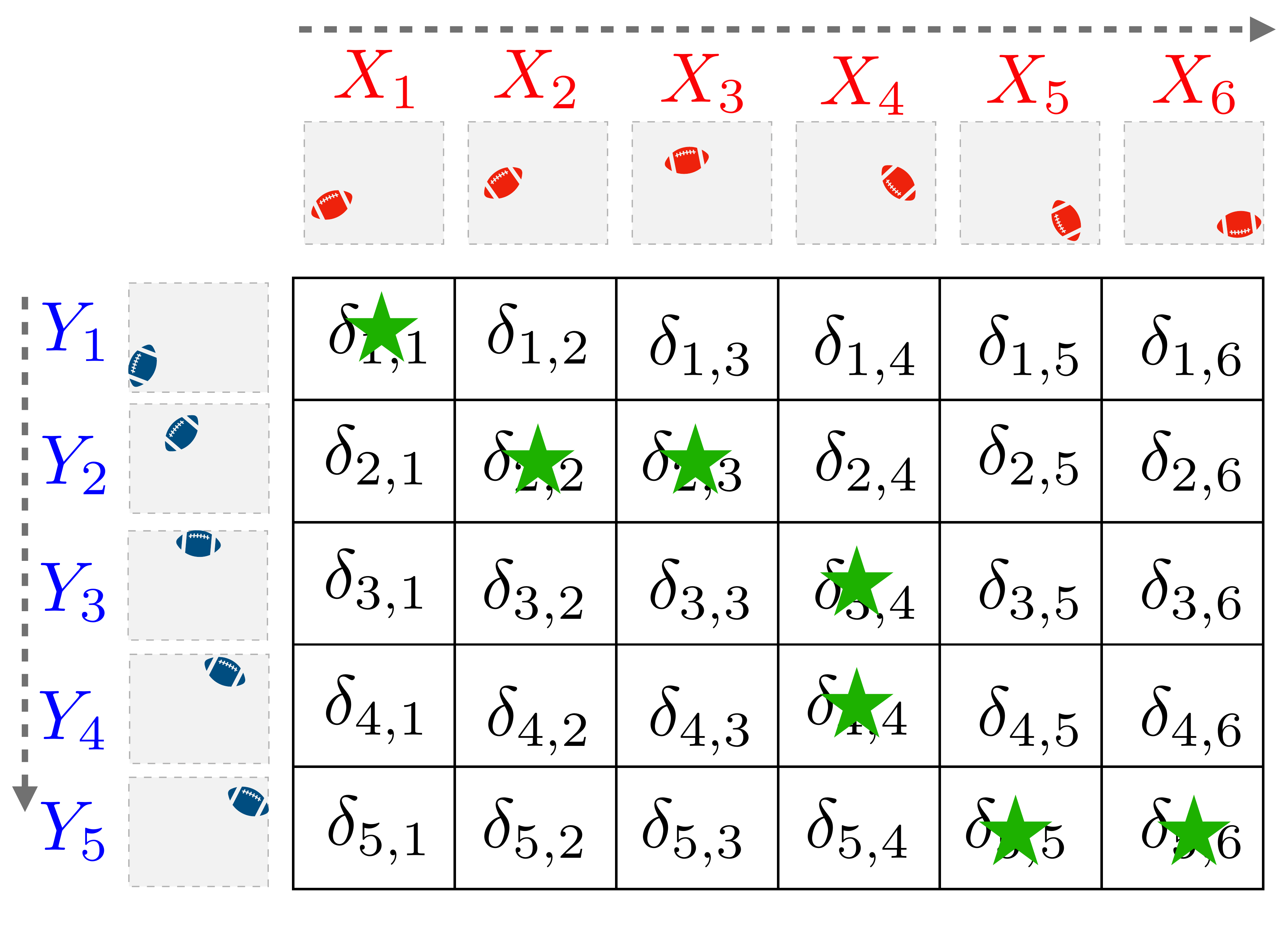}
	\captionof{figure}{Example of Dynamic time warping alignment between two time series of images given a pairwise distance matrix. \label{f:dtw-example}}
\end{minipage}
\hspace{0.2cm}
\begin{minipage}{0.61\linewidth}
	\hrule
	\vspace{0.1cm}
	\captionof{algorithm}{BP recursion to compute $\sdtw$ \citep{cuturi17} \label{a:dynamicprogram}}
	\begin{algorithmic}
		\vspace{-0.2cm}
		\hrule
		\vspace{0.1cm}
		\STATE {\bfseries Input:}  data $\bx, \by$ soft-min parameter $\beta$ and distance function $\delta$ 
		\STATE {\bfseries Output:} $\sdtw(\bx, \by) = r_{T_1, T_2}$
		\STATE $r_{0,0} = 0; r_{0,j} = r_{i, 0} =\infty$ for $i \in \intset{T_1}$, $j \in \intset{T_2}$ 
		\FOR{$i=1$ {\bfseries to} $T_1$}
		\FOR{$j=1$ {\bfseries to} $T_2$}
		\STATE $r_{i,j} = \delta(\bx_i, \by_j) + {\softmin}_\beta(r_{i-1, j-1}, r_{i-1, j}, r_{i, j-1})$
		\ENDFOR
		\ENDFOR
	\end{algorithmic}
	\hrule
\end{minipage}

\paragraph{Algorithmic differentiation}
When $\beta > 0$, differentiating \eqref{eq:sdtw} with respect to $\bx$ yields:
\begin{equation}
\label{eq:gradient-sdtw}
\nabla_{\bx} \sdtw(\bx, \by, \Delta) = \left(\frac{\partial \Delta(\bx, \by)}{\partial \bx}\right)^\top E_{\beta}(\bx, \by) \enspace ,
\end{equation}
where $E_{\beta}(\bx, \by) \eqdef \frac{\partial \sdtw}{\partial \Delta}(\bx, \by) = \frac{\sum_{\cA_{T_1, T_2}} e^{-\frac{\langle \bA, \Delta(\bx, \by)\rangle}{\beta}} A}
{\sum_{\cA_{T_1, T_2}} e^{-\frac{\langle \bA, \Delta(\bx, \by)\rangle}{\beta}}}$ can be interpreted as a weighted average alignment.
To compute $E_{\beta}(\bx, \by)$, \cite{cuturi17} proposed to back propagate the forward recursion of Algorithm \ref{a:dynamicprogram}, starting from $E_{T_1, T_2}$ down to $E_{0. 0}$. Indeed, the value of $\sdtw$ is stored in the last alignment cost $r_{T_1,T_2}$. Thus differentiating $\sdtw$ with respect to any  $r_{i, j}$ only involves the terms of 
$r_{i-1, j}, r_{i, j-1}$ and $r_{i-1, j-1}$. Differentiating the softmin operation of the forward pass yields the backward recursion of Algorithm \ref{a:backprop}.
\begin{algorithm}
	\caption{Backward recursion to differentiate sdtw \citep{cuturi17}.\label{a:backprop}}
	\begin{algorithmic}
		\STATE {\bfseries Input:} $\bx, \by$, parameter $\beta$, distance $\delta$ and intermediary alignment matrix $R$
		\STATE {\bfseries Output:} $E = E_\beta(\bx, \by)$
		\STATE $r_{i,m+1} = r_{n+1,j} = -\infty$, $i \in \intset{n}$, $j \in \intset{m}$ 
		\STATE $e_{i,m+1} = r_{n+1,j} = 0$, $i \in \intset{n}$, $j \in \intset{m}$ 
		\STATE $\delta_{i,m+1} = \delta_{n+1,j} =0$, $i \in \intset{n}$, $j \in \intset{m}$ 
		\STATE $\delta_{n+1,m+1} = 0, e_{n+1,m+1} =1, r_{n+1, m+1} = r_{n, m}$ 
		\FOR{$i=1$ {\bfseries to} $n$}
		\FOR{$j=1$ {\bfseries to} $m$}
		\STATE $a = \exp \frac{1}{\beta}(r_{i+1, j} - r_{i, j} - \delta_{i+1, j})$
		\STATE $b = \exp \frac{1}{\beta}(r_{i, j + 1} - r_{i, j} - \delta_{i, j + 1})$
		\STATE $c = \exp \frac{1}{\beta}(r_{i+1, j+1} - r_{i, j} - \delta_{i+1, j+1})$
		\STATE $e_{i,j} = a e_{i+1, j} + b e_{i, j+1} + c e_{i+1, j+1} $
		\ENDFOR
		\ENDFOR
	\end{algorithmic}
\end{algorithm}

\subsection{Delannoy numbers}
Delannoy numbers arise naturally when working with Dynamic time warping: they correspond to the number of feasible alignments $\cA_{T_1, T_2}$ and are usually denoted by $D(T_1 - 1, T_2 - 1)$~\citep{cuturi11}. The role played by Delannoy numbers is even more apparent when working with Soft-DTW which is only natural since the minimum is replaced with an exponential weighted sum over all feasible alignments.

For the sake of convenience, we consider the shifted Delannoy sequence starting at $n=m=1$ so that: $\card(\cA_{m, n}) = D_{m, n}$ for all integers $m, n \geq 1$. Formally, the Delannoy sequence can be defined recursively by:
\begin{definition}[Delannoy sequence]
	\label{def:delannoy}
	The Delannoy number $D_{m, n}$ corresponds to the number of paths from $(1, 1)$ to $(m, n)$ in a $(m \times n)$ lattice where only  $\rightarrow, \downarrow, \searrow$ movements are allowed. It can also be defined with the recursion $\forall m, n \in \bbN^{\star}$:
	\begin{align}
	& D_{1, n} = D_{m, 1} = 1 \\
	&  D_{m + 1, n + 1} = D_{m , n + 1} + D_{m + 1, n} + D_{m , n }  \label{eq:delannoy-rec} \enspace .
	\end{align}
\end{definition}
\citet{janati20-aistats} showed that there exists a positive constant $c$ such that the diagonal elements $D_{m, m}$ verify the bounded growth inequality $D_{m + 1, m + 1} \leq c^2 D_{m, m}$.
The following proposition provides tighter inequalities that bound the growth of the central Delannoy sequence from both ends. 
\begin{proposition}
	\label{prop:growth}
	Let $c = 1 + \sqrt{2}$ and $\sigma = \frac{21}{22}c^2 - 5$.
	The  central (diagonal) Delannoy sequence $D_m \eqdef D_{m, m}$ verifies:
	\begin{align}
	\frac{D_{m+1}}{D_m} \leq c^2 \frac{m}{m + \frac{1}{2}}  \quad \forall m \geq 1 \label{eq:growth-right}\\
	\frac{D_{m+1}}{D_m}  \geq  c^2 \frac{m}{m + \sigma} \quad \forall m \geq 5 \label{eq:growth-left}
	\end{align}
\end{proposition}
Reiterating this inequality over the course of the time series leads to the following corollary which is crucial to derive a simple heuristic to set the $\sdtw$ hyperparameter $\beta$ in practice.

\begin{corollary}
\label{cor:delannoy-bounds}
Let $T > m \geq 1$, $c = 1 + \sqrt{2}$ and $\sigma = \frac{21 c^2}{22} - 5 \,(\approxeq 0.56)$. The central Delannoy numbers verify:
\begin{align}
\label{eq:delannoy-sqrt}
\begin{split}
c^{2(T - m)}  \frac{D_m}{D_T} \geq \left(\frac{T}{m e} \right)^{\frac{1}{2}} \, \text{ for } m \geq 1  \\ 
c^{2(T - m)}  \frac{D_m}{D_T}\leq \left(\frac{T-1}{m-1} \right)^\sigma\, \text{ for } m \geq 5
\end{split}
\end{align}
\end{corollary}

\textsc{Proof of Proposition
\ref{prop:growth}.}
The central (or diagonal) Delannoy numbers $D_m$ verify the 2-stages recursion equation for any $m \geq 2$ ~\citep{stanley11}:
	\begin{equation}
	\label{eq:central-recursion}
	m D_{m + 1} = (6m - 3) D_{m} - (m - 1)D_{m-1}
	\end{equation}
We are going to prove both inequalities by induction. 
\paragraph{Inequality \eqref{eq:growth-right}}
For $m = 1$, we have $D_2=3 \leq 2 + \frac{4}{3}\sqrt{2} = \frac{2}{3} c^2= \frac{2}{3} c^2D_1$. Assume that \eqref{eq:growth-left} holds for some $m \geq 2$. 
From \eqref{eq:central-recursion} and the induction assumption:
\begin{align}
(m + 1)D_{m+2} &= (6m + 3)D_{m + 1} - mD_{m} \\
& \leq (6m + 3)D_{m + 1} - m \frac{m + \frac{1}{2}}{c^2 m} D_{m+1} \\
& \leq (6m + 3 - \frac{m + \frac{1}{2}}{c^2}) D_{m + 1} \\
& = \frac{(6c^2 - 1)m + \frac{6c^2 - 1}{2}}{c^2}D_{m + 1} \\
&= c^2 (m +  \frac{1}{2}) D_{m + 1}
\end{align} 
where we used the fact that $1 / c^2 = \frac{1}{3 + 2\sqrt{2}} = 3 - 2\sqrt{2} = 6 - c^2$, hence $6c^2 - 1= c^4$. Therefore: $$\frac{D_{m+2}}{D_{m+1}} \leq c^2 \frac{m +  \frac{1}{2}}{m+1} \enspace.$$ To conclude, it suffices to show that for all $m\geq 5$:
\begin{align}
\frac{m + \frac{1}{2}}{m + 1} \leq \frac{m + 1}{m + \frac{3}{2}}  
\end{align}  
which is equivalent to:
\begin{align}
& \left(m+ \frac{1}{2}\right)\left(m  + \frac{3}{2}\right) \leq (m + 1)^2 \\
& \Leftrightarrow m^2 + 2m + \frac{3}{4} \leq m^2 + 2m + 1 \\
& \Leftrightarrow  \frac{3}{4} \leq 1
\end{align} 
\paragraph{Inequality \eqref{eq:growth-left}}
For $m = 5$, we have with numerical evaluation $\frac{D_6}{D_5} - c^2 \frac{5}{5 + \sigma} \geq 0$. Assume that \eqref{eq:growth-left} holds for some $m \geq 6$. 
From \eqref{eq:central-recursion} and the induction assumption:
\begin{align}
(m + 1)D_{m+2} &= (6m + 3)D_{m + 1} - mD_{m} \\
& \geq (6m + 3)D_{m + 1} - m \frac{m + \sigma}{c^2 m} D_{m+1} \\
& = (6m + 3 - \frac{m + \sigma}{c^2}) D_{m + 1} \\
& = \frac{(6c^2 - 1)m + 3c^2 - \sigma}{c^2}D_{m + 1} \\
&= c^2 (m + \frac{3c^2 - \sigma}{c^4}) D_{m+1}
\end{align} 
where we used the fact that $1 / c^2 = \frac{1}{3 + 2\sqrt{2}} = 3 - 2\sqrt{2} = 6 - c^2$, hence $6c^2 - 1= c^4$.  Therefore: $$\frac{D_{m+2}}{D_{m+1}} \geq c^2 \frac{m + \frac{3c^2 - \sigma}{c^4}}{m+1} \enspace.$$ To conclude, it suffices to show that for all $m\geq 2$:
\begin{align}
\frac{m + \frac{3c^2 - \sigma}{c^4}}{m + 1} \geq \frac{m + 1}{m + \sigma + 1}  
\end{align}  
which is equivalent to:
\begin{align}
& \left(m+ \frac{3c^2 - \sigma}{c^4}\right)\left(m  + \sigma + 1\right) \geq (m + 1)^2 \nonumber \\
& \Leftrightarrow \left( \frac{3c^2 - \sigma}{c^4} + \sigma - 1\right)  m + \frac{3c^2 - \sigma}{c^4} (\sigma + 1) - 1\geq 0 \label{eq:proof-tau}
\end{align} 
Numerical evaluation shows that  $\frac{3c^2 - \sigma}{c^4} + \sigma - 1 \geq 0.06$  and 
that $\frac{3c^2 - \sigma}{c^4} (\sigma + 1) - 1 \geq - 0.24$. Thus \eqref{eq:proof-tau} is verified for $m  \geq 5$.  $\blacksquare$

\textsc{Proof of Corollary \ref{cor:delannoy-bounds}.} Combining both inequalities of proposition \ref{prop:growth} leads to:
\begin{align*}
&\prod_{k=m}^{T-1}  c^2 \frac{k}{k + \sigma} \leq \frac{D_T}{D_m} \leq \prod_{k=m}^{T-1}  c^2 \frac{k}{k + \frac{1}{2}}  \\ 
\Leftrightarrow &\prod_{k=m}^{T-1}  \frac{k + \frac{1}{2}}{k} \leq c^{2(T - m)} \frac{D_m}{D_T} \leq \prod_{k=m}^{T-1}  \frac{k + \sigma}{k}  \\ 
\Leftrightarrow &\prod_{k=m}^{T-1}  \frac{k + \frac{1}{2}}{k} \leq c^{2(T - m)} \frac{D_m}{D_T} \leq \prod_{k=m}^{T-1}  \frac{k + \sigma}{k}  \\
\Leftrightarrow &\exp\left[\sum_{k=m}^{T-1}  \log\left(1 + \frac{1}{2k}\right)\right] \leq c^{2(T - m)} \frac{D_m}{D_T} \leq \exp\left[\sum_{k=m}^{T-1}  \log\left(1 + \frac{\sigma}{k}\right)\right] 
\end{align*}
Let $z \in [\frac{1}{2}, 1[$. Using the inequalities $\frac{x}{1+x} \leq \log(1 + x) \leq x$ which holds for $x \geq -1$:
\begin{align*}
&\sum_{k=m}^{T-1} \frac{z}{z + k}  \leq \sum_{k=m}^{T-1} \log\left(1 + \frac{z}{k}\right) \leq \sum_{k=m}^{T-1} \frac{z}{k} \\
\Rightarrow & z \sum_{k=m}^{T-1} \frac{1}{1 + k}  \leq \sum_{k=m}^{T-1} \log\left(1 + \frac{z}{k}\right) \leq z\sum_{k=m}^{T-1} \frac{1}{k} \\
\Leftrightarrow & z \left(\sum_{k=0}^{T-1} \frac{1}{1 + k}  - \sum_{k=0}^{m-1} \frac{1}{1 + k} \right) \leq \sum_{k=m}^{T-1} \log\left(1 + \frac{z}{k}\right) \leq z\left(\sum_{k=1}^{T-1} \frac{1}{k}  - \sum_{k=1}^{m-1} \frac{1}{k} \right)\\
\Leftrightarrow & z \left(\sum_{k=1}^{T} \frac{1}{k}  - \sum_{k=1}^{m} \frac{1}{k} \right) \leq \sum_{k=m}^{T-1} \log\left(1 + \frac{z}{k}\right) \leq z\left(\sum_{k=1}^{T-1} \frac{1}{k}  - \sum_{k=1}^{m-1} \frac{1}{k} \right)\\
\end{align*}
Finally, using the classical bounds of the Harmonic series~\citep{chen03}:
\begin{equation}
\log(n) + \gamma + \frac{1}{2n + 1} \leq \sum_{i=1}^n \frac{1}{i} \leq \log(n) + \gamma + \frac{1}{2n - 1} \enspace,
\end{equation}
it holds:
\begin{align*}
\frac{1}{z} \sum_{k=m}^{T-1} \log\left(1 + \frac{z}{k}\right) &\leq \log(T-1) + \frac{1}{2T - 2} - \log(m-1) -  \frac{1}{2m - 1} \\
 &\leq \log(T) + \frac{1}{2T - 2} - \log(m-1) -  \frac{1}{2m - 1} \\
 &\leq \log\left(\frac{T - 1}{m -1}\right) + \underbrace{\frac{1}{2}\frac{2(m - T) + 1}{(T-1)(2m - 1)}}_{<0} \\
 &\leq \log\left(\frac{T - 1}{m -1}\right) 
\end{align*}
and similarly:
\begin{align*}
\frac{1}{z}\sum_{k=m}^{T-1} \log\left(1 + \frac{z}{k}\right) &\geq \log(T) + \frac{1}{2T + 1} - \log(m) -  \frac{1}{2m + 2} \\
&\geq \log\left( \frac{T}{m}\right) - 1
\end{align*}
Taking the exponential after substituting $z$ by $\sigma$ (resp. $\frac{1}{2}$) provides the upper (resp. lower) bound.  $\blacksquare$.

We conclude this section on Delannoy numbers by restating a result of \citep{janati20-aistats} which will be useful later on. It provides bounds on the off-diagonal Delannoy numbers which lead to the quadratic lower bound of Soft-DTW.
\begin{proposition}
		Let $c = 1 + \sqrt{2}$.  $\forall m, i \in \bbN^\star$:
	\begin{align}
	\label{eq:delannoy-ineq}
	&D_{m, m + i}  \leq c \Phi_{m, i} D_{m, m + i - 1} \\ 
	&c \Psi_{m, i} D_{m, m + i}  \leq D_{m + 1, m + i}
	\end{align}
	Where
	\[
	\left\{	 \begin{array}{l} 
	\Phi_{m, i} = 1 - \frac{ (1 - \frac{1}{c}) (i - 1) + \frac{1}{c}}{m + i - 1} \\
	\Psi_{m, i} = 1 + \frac{(1 - \frac{1}{c}) (i - 1)}{m} 
	\end{array}
	\right.
	\]
	Moreover, combining the inequalities above with a running $i=1..k$ we get
	for any $m, m', k \in \bbN^\star$ such that $m + m' \leq T - 1$ and $ k \leq \min(T - m, m' - 1)$:
	\label{prop:quad-lb}
	\begin{equation}
	\label{eq:quad-lb}
	\log \left( \frac{ D_{m, m}  D_{m' , m'} }{ D_{m + k, m}  D_{m' - k , m'} } \right)  \geq P(k)
	\end{equation}
	where $P(k) = \alpha k(k-1) +  \rho k  + \frac{1}{3T}$ with  $\alpha = \frac{2 - \sqrt{2}}{T} > 0$ and $\rho = \frac{3\sqrt{2} -4 }{3T} > 0$.
\end{proposition}

\subsection{Time sensitivity of Soft-DTW and effect of $\beta$}
In \citep{janati20-aistats}, Soft-DTW was shown to increase quadratically with the size of the temporal shift between two time series. Formally let $\bx, \by \in \bbR^{T}$ such that $\by$ is temporally shifted w.r.t. $\bx$ by k time steps : 
\begin{align}
\by_{i + k} = \bx_i \quad \forall i \in \intset{1, T - k} 
\end{align}
Then the following holds:
\begin{proposition}
	\label{prop:lb}
	Let $r = \min_{i,j}\{\Delta(\bx, \bx)_{ij} | \Delta(\bx, \bx)_{ij} > 0\}$.  Denote $ m = \argmin_{i \in \intset{1, T - 1}} \{\bx_{i+1} \neq \bx_i\} $	 $m' = T- \argmax_{i \in \intset{1, T - 1}} \{\bx_{i+1} \neq \bx_i\}  $.
	 
	 If $0 < \beta \leq \frac{r}{\log(3T D_{T, T})}$ :
	\begin{align}
	\label{eq:lb}
	\sdtw(\bx, \by) -  \sdtw(\bx, \bx)  \geq
	\beta \log \left( \frac{ D_{m, m}  D_{m' , m'} }{ D_{m + k, m}  D_{m' - k , m'} } \right) - \frac{\beta}{3T}
	\end{align}
\end{proposition}
\proof see \citep{janati20-aistats} for a proof.

\begin{figure}[t]
	\begin{minipage}{0.45\linewidth}
		\includegraphics[width=\linewidth]{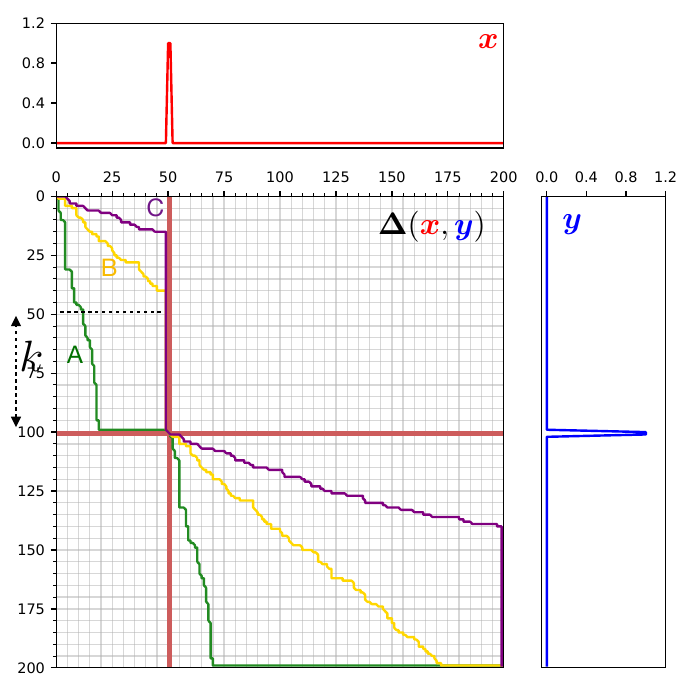}
	\end{minipage}
	\begin{minipage}{0.5\linewidth}
		\includegraphics[width=\linewidth]{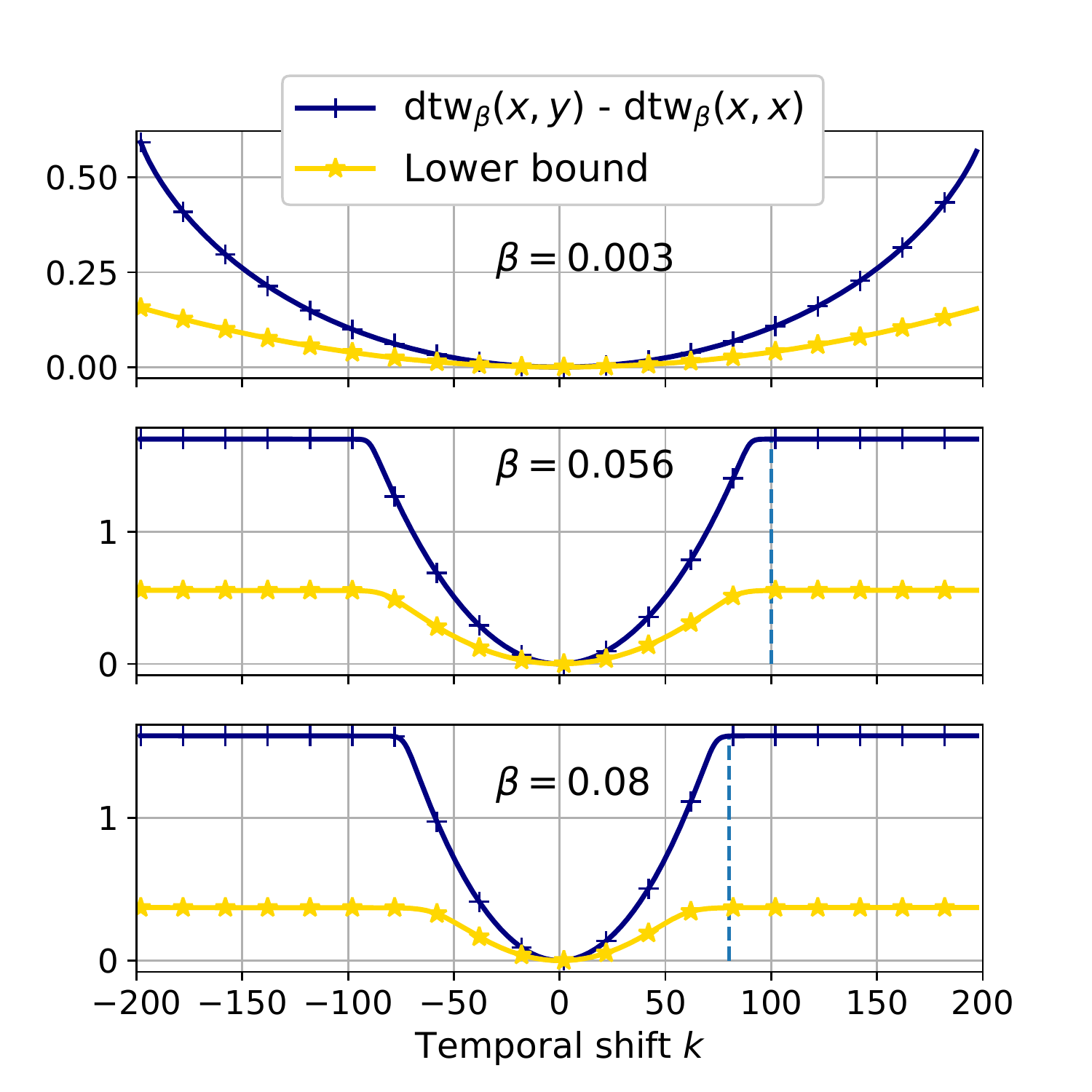}
	\end{minipage}
	\caption{Left: example of 3 DTW alignment paths (A, B and C) between $\bx$ and $
		\by$ with a temporal 50-shift. The heatmap of the distance matrix $\Delta$ (here squared Euclidean) shows 2 red bars where the distance is not equal to 0 except at their intersection. An alignment path has 0 cost if and only if it does not cross the red lines. Compared to the figure of \citep{janati20-aistats}, here we consider the simpler case of Dirac time series. This setting allows one to obtain a bound for all values of $\beta$, including the asymptotes for large temporal shifts displayed on the right side figure above.
		\label{f:bound}}
\end{figure}

An example of a temporal 50-shift is illustrated in Figure~\ref{f:bound}. The heatmap of the squared Euclidean cost matrix $\Delta$ shows two white rectangles where all alignments A, B and C have the same cost of 0. Since $\dtw$ is defined as the minimum of all alignment costs, all these paths are equivalent. Shifting $\by$ temporally would move the red horizontal line downwards, changing the set of alignments with cost 0 without affecting the $\dtw$ value (0). However, when $\beta>0$, $\sdtw$ computes a weighted sum of all possible paths, which is affected by temporal shifts by including the number of equivalent paths. This number of equivalent paths is expressed as the product of Delannoy numbers as showed by proposition \ref{prop:lb} which, combined with proposition \ref{prop:quad-lb} provides a quadratic lower bound for temporal shifts.

\paragraph{A tighter lower bound}
While the previously considered time series covered a wide range of scenarios, the obtained result requires $\beta$ to be too small, thereby not providing any insight on how $\sdtw$ behaves when $\beta$ increases. In the following paragraph, we relax this assumption on $\beta$ in order to find a tighter lower bound than the one given in \eqref{eq:lb}. We consider the simplified setting of Dirac univariate time series $\bx, \by$ such that $\by$ is ahead of $\bx$ by k time steps (see Figure \ref{f:dtw-example}). Formally, let $\bx, \by \in
\bbR^T$ such that for some $t^\star \in \intset{1, T}$ and $1 \leq k\leq T - t^\star$:
\begin{align}
\begin{split}
\label{eq:dirac}
t \neq t^\star \Rightarrow \bx_t = 0 \\
t \neq t^\star + k \Rightarrow \by_t = 0 \\
\bx_{t^\star} = \by_{t^\star + k} = c \in \bbR
\end{split}
\end{align}
This simplified setting allows for tighter bounds of Soft-DTW. 
\begin{proposition}
\label{prop:dirac-bound}
	Consider $\bx$ and $\by$ as defined in \eqref{eq:dirac}. Let $r = \Delta(c, 0)$,  $T  \geq 6$, $c = 1 + \sqrt{2}$ and $\sigma = \frac{21 c^2}{22} - 5 \,(\approxeq 0.56)$.Then:
\begin{align}
\label{eq:dirac-bound}
\sdtw(\bx, \by) -  \sdtw(\bx, \bx)  \geq  - \beta\log\left( e^{- P(k)} (1 - \lambda_{\beta}) + \lambda_{\beta} H\right)
\end{align}
where: $\lambda_{\beta} = e^{-\frac{r}{\beta}}$, $H = 92 T^\sigma$ and $P$ is the quadratic bound defined in Proposition \ref{prop:quad-lb}.
\end{proposition}
\proof First let's upper bound $\sdtw(\bx, \by)$. Notice that since $\bx, \by$ are Dirac time series, the elements of the distance matrix $\Delta(\bx, \by)$ are either equal to 0 or $r$. Therefore, the cost of any path A given by $\langle A, \Delta(\bx, \by)\rangle$ can be written as $qr$ for some $q \in \bbN$. More specifically, $q$ corresponds to the number of times the path $A$ meets the non-zero elements of $\Delta(\bx, \by)$. Therefore, denoting the number of feasible alignments corresponding to each $q$ by $M_q(\bx, \by)$ it holds:
\begin{equation}
\label{eq:sdtw-dirac-q}
\sum_{A \in \cA_{T, T}} e^{-\frac{\langle A, \Delta(\bx, \by)\rangle}{\beta}} = \sum_{q=0}^{T + k} M_q(\bx, \by) e^{-\frac{qr}{\beta} } = M_0(\bx, \by) + e^{-\frac{r}{\beta}} \sum_{q=1}^{T + k} M_k(\bx, \by) e^{-\frac{r(q - 1)}{\beta}} 
\end{equation}
Therefore: 
\begin{align*}
\sum_{A \in \cA_{T, T}} e^{-\frac{\langle A, \Delta(\bx, \by)\rangle}{\beta}}
 &\leq M_0(\bx, \by) + e^{-\frac{r}{\beta}} \sum_{q=1}^{T + k} M_k(\bx, \by)\\
 &= M_0(\bx, \by) + e^{-\frac{r}{\beta}} (D_{T} - M_0(\bx, \by)) \\
&= (1-\lambda_{\beta}) M_0(\bx, \by) + \lambda_{\beta} D_T
\end{align*}
and similarly:
\begin{align*}
\sum_{A \in \cA_{T, T}} e^{-\frac{\langle A, \Delta(\bx, \bx)\rangle}{\beta}}  &\geq M_0(\bx, \bx) 
\end{align*}
where $M_0(\bx, \by) = D_{t^\star - 1, t^\star - 1 + k} D_{T - t^\star, T - t^\star - k}$ and $M_0(\bx, \bx) = D_{t^\star - 1} D_{T - t^\star}$. Therefore, combining both inequalities after introducing $-\log$ leads to:
\begin{align*}
\sdtw(\bx, \by) -  \sdtw(\bx, \bx)  &\geq  - \beta\log\left( \frac{ M_0(\bx, \by) (1 - \lambda_{\beta}) + \lambda_{\beta} D_T}{M_0(\bx, \bx)}\right) \\
&=  - \beta\log\left( \frac{ D_{t^\star - 1, t^\star - 1 + k} D_{T - t^\star, T - t^\star - k} }{D_{t^\star - 1} D_{T - t^\star}} (1 - \lambda_{\beta}) + \lambda_{\beta} \frac{D_T}{D_{t^\star - 1} D_{T - t^\star}}\right)
\end{align*}
On one hand applying Proposition \ref{prop:quad-lb} with $m = t^\star - 1$ and $m' = T - m - 1$ provides:
\begin{equation*}
\frac{ D_{t^\star - 1, t^\star - 1 + k} D_{T - t^\star, T - t^\star - k} }{D_{t^\star - 1} D_{T - t^\star}} \leq e^{-P(k)}
\end{equation*}
And on the other, using Corollary \ref{cor:delannoy-bounds} we can get the H upper bound. For $t^\star \geq 1$:
\begin{equation*}
\frac{D_T}{D_{T-t^\star}} \leq \sqrt{\frac{(t^\star - 1)e}{T}} \frac{1}{c^{2(T - t^\star + 1)}} \leq \sqrt{\frac{t^\star e}{T}} c^{2t^\star}
\end{equation*}
and if $t^\star -1 \geq 5$: 
\begin{equation*}
\frac{D_5}{D_{t^\star - 1}} \leq \left(\frac{T - t^\star - 1}{4}\right)^{\sigma} c^{2(T - t^\star - 5)} \leq \left(\frac{T - t^\star}{4}\right)^{\sigma} \frac{1}{c^{2(t^\star - 6)}}
\end{equation*}
Combining the two leads to:
\begin{equation*}
\frac{D_T}{D_{t^\star - 1} D_{T - t^\star}} \leq \frac{c^{12}}{D_5} \left(\frac{T - t^\star}{4}\right)^{\sigma} \sqrt{\frac{t^\star e}{T}}
\end{equation*}
Maximizing the upper bound with respect to $t^\star$ leads to the maximizer $t^\star = \frac{T}{2 \sigma + 1}$. Substituting shows that:
$$ (T - t^\star)^{\sigma} \sqrt{\frac{t^\star }{T}} \leq  \left(T \frac{2\sigma}{1 + 2 \sigma} \right)^{\sigma} \sqrt{\frac{1 }{1 + 2\sigma}}  $$
Finally, numerical evaluation gets rid of the constants: 
$$ \frac{D_T}{D_{t^\star - 1} D_{T - t^\star}} \leq H$$
If however $t^\star \leq 5$: 
$$ \frac{D_T}{D_{t^\star - 1} D_{T - t^\star}}  \leq  \sqrt{\frac{t^\star e}{T}} \left(\frac{c^2}{3}\right)^{t^\star}  \leq \sqrt{e} \left(\frac{c^2}{3}\right)^{5}  \approx 45.63 < 92 < H $$

Multiplying by  $1 - \lambda_{\beta} \geq 0$, adding $\lambda_{\beta} H$ and applying $-\log$  ends the proof $\blacksquare$.
\paragraph{Effect of $\beta$} When $\beta $ is small enough, $\lambda_{\beta}$ goes to 0, thus the lower bound of Proposition \ref{prop:dirac-bound} can be very well approximated by the quadratic bound $P(k)$. However when $\beta$ increases, $\lambda_{\beta} H$ increases which will dominate the log argument for a sufficiently large $k$. Using the example of Figure \ref{f:dtw-example}, we compute both sides of Equation \eqref{eq:dirac-bound} for 3 values of $\beta$. Figure \ref{f:bound} shows that $\sdtw$ saturates for a certain temporal shift $k_{\max}$  beyond which it is no longer sensitive to temporal lags. This phase transition is also observed by the lower bound \eqref{eq:dirac-bound}. This provides a heuristic to set $\beta$ based on a predefined $k_{\max}$ i.e the largest temporal shift the user is willing to capture. Notice that such a point does not always exist (when $\beta$ is too small) as it may be larger than the time series length T (see top example, Figure \ref{f:bound}). 
%
\begin{proposition}
	\label{prop:heuristic-beta}
	Let $T \geq 6$, $1 \leq k_{\max}$ and $0 < \eta < 1$. Using the same notations of Proposition \ref{prop:dirac-bound}, define the lower bound function: $$ \LB_{\beta}: k \mapsto - \beta\log\left( e^{- P(k)} (1 - \lambda_{\beta}) + \lambda_{\beta} H\right) $$
	If $\beta \geq \frac{r}{P(k_{\max}) + \log\left( (e^\eta - 1)H \right)} $  then:
	\begin{equation}
	\label{eq:heuristic}
	0\leq \frac{\underset{k\to +\infty}{\text{lim}} \LB_{\beta}(k) - \LB_{\beta}(k_{\max})  }{\beta} \leq \eta
	\end{equation}
\end{proposition}
\proof
It is straightforward to see that $\underset{k\to +\infty}{\text{lim}} \LB_{\beta}(k) = -\beta\log(\lambda_{\beta} H)$. Therefore on one hand:
\begin{align*}
	&\frac{\underset{k\to +\infty}{\lim} \LB_{\beta}(k) - \LB_{\beta}(k_{\max})  }{\beta} \leq \eta \\
	\Leftrightarrow  & \log\left( e^{- P(k)} (1 - \lambda_{\beta}) + \lambda_{\beta} H\right)  - \log(\lambda_{\beta} H) \leq \eta \\
	\Leftrightarrow & e^{- P(k_\max)} (1 - \lambda_{\beta}) + \lambda_{\beta} H  \leq e^{\eta}  \lambda_{\beta} H \\
	\Leftrightarrow & \lambda_\beta \geq \frac{e^{- P(k_\max)}}{ (e^{\eta} - 1) H + e^{-P(k_\max)}} \\
\end{align*}
On the other hand:
\begin{align*}
&\beta \geq \frac{r}{P(k_{\max}) + \log\left( (e^\eta - 1)H \right)} \\
\Rightarrow & -\frac{r}{\beta} \geq -P(k_{\max}) - \log\left( (e^\eta - 1)H \right) \\
\Rightarrow& \lambda_\beta \geq \frac{e^{-P(k_\max)}}{(e^\eta - 1)H } \geq \frac{e^{- P(k_\max)}}{ (e^{\eta} - 1) H + e^{-P(k_\max)}} 
\end{align*} 
Thus the upper bound in \eqref{eq:heuristic} holds. The lower bound follows from the positivity of $e^{- P(k)} (1 - \lambda_{\beta}) \, \blacksquare$.

 Proposition \ref{prop:heuristic-beta} provides a sufficient condition to set $\beta$ such that the lower bound LB saturates for a certain $k_{\max}$. In the examples shown in Figure \ref{f:bound}, $\beta$ was set using this heuristic with $\eta = 0.01$ and $k_{\max} \in \{500, 100, 80\}$ respectively top to bottom. The dotted vertical lines highlight the choice of $k_{\max}$ which is very close to the saturation point of $\sdtw$. In practice, we use the same heuristic by setting $r = \max_{i, j} {\Delta(\bx_i, \by_j)}$.

\section{Soft-DTW barycenters via alternating optimization}
\label{s:ao}

Consider a dataset with N multivariate time series $\bx_1, \dots, \bx_N$ assumed to have the same dimension $p$ and respective time lengths $T_1, \dots, T_N$. Let $w_1, \dots, w_n$ be a set of positive weights summing to one. The Soft-DTW barycenter with cost $\Delta$ and fixed length $T$ is defined as:
	\begin{align}
	\label{eq:barycenter}
	\bar{\bx} =& \argmin_{\bx \in \bbR^{p, T}} \sum_{i=1}^N w_i \sdtw(\bx_i, \bx) \\ =& \argmin_{\bx \in \bbR^{p, T}} -\sum_{i=1}^N w_i \beta \log\left( \sum_{A \in \cA_{T, T_i}} e^{-\frac{\langle A, \Delta(\bx, \bx_i)\rangle}{\beta}} \right)
	\end{align}

\paragraph{Alternating optimization}
Provided that $\Delta$ is differentiable, the most straightforward solution to \eqref{eq:barycenter} would probably be to use a Quasi-Newton method. However, since we intend to use an OT loss for $\Delta$, computing each gradient step would require $T\sum_{i=1}^N T_i$ OT gradients. Instead, we use Fenchel duality to obtain an alternating optimization problem that non only avoids the computation of the gradients of $\Delta$ but also spares us any form of step-size backtracking. This is given by proposition \ref{prop:fenchel}.
\begin{proposition}
\label{prop:fenchel}
For the sake of convenience, provide the sets of binary matrices $\cA_{T_i, T}$ with some arbitrary indexation $\cA_{T_i, T} = \{A^1_i, \dots, A^{D_{T_i, T}}_i\}$ and let $S_K$ denote the probability simplex of $\bbR^K$. 
The Soft-DTW problem \eqref{eq:barycenter} is equivalent to the optimization problem:
\begin{equation}
\label{eq:fenchel}
	\min_{\bx \in \bbR^{p, T}} \min_{\substack{\theta_1 \in S_{D_{T_1, T}}  \\ \dots \\ \theta_N \in S_{D_{T_N, T} }}}  \sum_{i=1}^N\langle  \sum_{k=1}^{D_{T_k, T}}  w_i \theta^k_i A^k_i, \Delta(\bx_i, \bx) \rangle   + \beta H(\theta_i)
\end{equation}
\end{proposition}
\proof A standard result in convex optimization theory states that the Fenchel conjugate of entropy is logsumexp. Formally, for $\bx \in \bbR^K$:
\begin{align}
(\beta H)^{\star}(\bx) \eqdef  \max_{\theta \in S_K} \langle \bx, \theta\rangle - \beta H(\theta)
								= \beta \log\left(\sum_{k=1}^K e^{-\frac{\bx_k}{\beta}}\right)
\end{align}
Thus:
\begin{equation}
- \beta \log\left(\sum_{k=1}^K e^{-\frac{\bx_k}{\beta}}\right)  = \min_{\theta \in S_K} \langle -\bx, \theta\rangle + \beta H(\theta).
\end{equation}
Therefore, the barycenter loss \eqref{eq:barycenter} can be written:
\begin{align}
	 \min_{\bx \in \bbR^{p, T}} &\sum_{i=1}^N w_i \sdtw(\bx_i, \bx)  = \min_{\bx \in \bbR^{p, T}} \sum_{i=1}^N - w_i \beta \log\left( \sum_{A \in \cA_{T, T_i}} e^{-\frac{\langle A, \Delta(\bx, \bx_i)\rangle}{\beta}} \right) \\
	&= \min_{\bx \in \bbR^{p, T}} \sum_{i=1}^N w_i \min_{\theta_i \in S_{D_{T_i, T}}}  \langle  \theta_i, \left(\langle A_i^1, \Delta(\bx_i, \bx) \rangle, \dots,  \langle {A^{D_{T_i, T}}}_i, \Delta(\bx_i, \bx) \rangle\right) \rangle   + \beta H(\theta_i) \\
	&= 	\min_{\bx \in \bbR^{p, T}} \sum_{i=1}^N w_i \min_{\theta_i \in S_{D_{T_i, T}}} \sum_{k=1}^{D_{T_k, T}} \langle \theta^k_i A^k_i, \Delta(\bx_i, \bx) \rangle   + \beta H(\theta_i) \\
	&=	\min_{\bx \in \bbR^{p, T}} \min_{\substack{\theta_1 \in S_{D_{T_1, T}}  \\ \dots \\ \theta_N \in S_{D_{T_N, T} }}}  \sum_{i=1}^N w_i \left( \left\langle  \sum_{k=1}^{D_{T_k, T}}  \theta^k_i A^k_i, \Delta(\bx_i, \bx) \right\rangle   + \beta H(\theta_i)\right)\enspace,
\end{align}
where the last equality follows from the separability of the sum with respective to the $\theta_i$ $\blacksquare$

The major benefit of the dual formulation of Proposition \eqref{prop:fenchel} is the ability to compute Fréchet means of $\Delta$ directly. This will be in particular crucial when we define $\Delta$ as an OT divergence for which Fréchet means are orders of magnitude faster to compute using Sinkhorn's algorithm than via gradient based methods~\citep{cuturi2018}.  While minimizing with respect to the $\theta_i$ seems computationally unfeasible due their large dimension, their update is actually not required to compute the new $\bx$. Instead, one needs to update the matrices $\bZ_i \eqdef \sum_{k=1}^{D_{T_k, T}}  w_i \theta^k_i A^k_i$ which are \emph{exactly} given by the gradients $\frac{\partial \sdtw(\bx_i, \bx)}{\partial \Delta} (\bx_i, \bx)$. Indeed, given that the loss is convex in $\theta_i$,  for a fixed $\bx$, the optimal $\theta_i$ verifies the KKT conditions for some Lagrange multiplier $\lambda_i$:
\begin{align*}
\left\{ \begin{array}{ll}
  \langle  A^k_i, \Delta(\bx_i, \bx) \rangle   + \beta \log(\theta_i^k) - \lambda_i = 0  \\
 \sum_{k=1}^{D_{T_i, T}}\theta_i^k = 1
 \end{array}
 \right.\end{align*}
 which leads to:
\begin{equation}
\theta_i^{k} =   \frac{e^{-\frac{\langle A^k_i, \Delta(\bx_i, \bx)\rangle}{\beta}}}{\sum_{k=1}^{D_{T_k, T}}  e^{ -\frac{\langle A^k_i, \Delta(\bx_i, \bx)\rangle}{\beta}}}
\end{equation}
Thus:
\begin{equation}
\bZ^i \eqdef \sum_{k=1}^{D_{T_k, T}}  w_i \theta^k_i A^k_i =   \frac{ \sum_{k=1}e^{-\frac{\langle A^k_i, \Delta(\bx_i, \bx)\rangle}{\beta}} A_i^k}{\sum_{k=1}^{D_{T_k, T}}  e^{ -\frac{\langle A^k_i, \Delta(\bx_i, \bx)\rangle}{\beta}}} = \frac{\partial \sdtw(\bx_i, \bx)}{\partial \Delta} (\bx_i, \bx) \enspace,
\end{equation}
which can be computed using Algorithm \ref{a:backprop}. Notice that to update $\bx$ computing the $\theta_i$ is not necessary, it is sufficient to update the $\bZ_i$. This leads to algorithm \ref{a:barycenter}.
\paragraph{Initialization} It is important to keep in mind the loss \eqref{eq:fenchel} is not jointly convex in $\bx$ and $\theta$. Thus, algorithm \ref{a:barycenter} is not guaranteed to converge to a global minimum. Nevertheless, in our experiments, initializing $\bx$ with a uniform distribution leads to meaningful barycenters with the desired spatio-temporal
properties.
 
 \begin{algorithm}
	\caption{Soft-DTW barycenter.\label{a:barycenter}}
	\begin{algorithmic}
		\STATE {\bfseries Input:} $\bx_1, \dots, \bx_N$, $\bx_0$, weights $w_1, \dots, w_N$, parameter $\beta$
		\STATE {\bfseries Output:}  solution of \eqref{eq:fenchel}
		\STATE {\bfseries Initialize $\bx = \bx_0 \in \bbR^{p, T}$, compute $\Delta(\bx_i, \bx)$ for all $i=1..N$ }
		\WHILE{not converged} 
		\FOR{$i=1$ {\bfseries to} $N$}
			\STATE{ Compute $\bZ^i$ with Algorithm \ref{a:backprop} }
		\ENDFOR
		\FOR{$t=1$ {\bfseries to} $T$}
			\STATE{ $\bx^t =  \argmin_{\ba \in \bbR^p} \sum_{i=1}^N \sum_{t'=1}^{T_i} w_i\bZ_{t',t}^i \Delta(\bx_i^{t'}, \ba)$}
		\ENDFOR
		\ENDWHILE
	\end{algorithmic}
\end{algorithm}

\section{Spatio-temporal barycenters}
\label{s:ot}

\subsection{$\widetilde{\uot}$ as debiased unbalanced OT}
We start this section by explaining how we use optimal transport to define the $\sdtw$ cost $\widetilde{\uot}$ between two non-negative  measures $\bx, \by$ with a fixed support given by $\cA = \{a_1, \dots, a_p\} \subset \bbR^d$. Since the support is fixed, $\bx, \by$ can be identified with vectors of non-negative weights i.e $\bx, \by \in \bbR_+^p$.

\paragraph{Entropy regularized unbalanced OT and Sinkhorn's algorithm} Let $\bC \in \bbR_+^{d, d}$ be the pairwise distance matrix given by $\bC_{ij} = c(a_i, a_j)$, where $c$ is a symmetric Lipschitz cost function such that  $\exp(- \frac{\bC}{\varepsilon})$ is positive semi-definite for any $\varepsilon > 0$. The matrix $\bC$ -- known as the ground metric --  defines the geometry that OT distances lift to compute transportation costs. Formally,  transporting a fraction of mass $\bP_{ij}$ from $a_i$ to $a_j$ is given by $\bP_{ij} \bC_{ij}$, the total cost of transport is given by $\langle \bP, \bC\rangle = \sum_{ij} \bP_{ij} \bC_{ij}$. To guarantee mass transportation, \cite{liero16} introduced the following formulation of unbalanced OT:
\begin{equation}
\label{eq:uot-savare}
\min_{\bP \in {\bbR_+}^{p\times p}} \, \langle \bP, \bC\rangle + \gamma \kl(\bP\mathds 1 | \bx) + \gamma \kl(\bP^\top \mathds 1 | \by) \enspace,
\end{equation}
where $\gamma >0$ is a fixed hyperparameter.  \citet{chizat17} generalized the above problem to other divergences than $\kl$ while adding entropy regularization \citep{cuturi13}, leading to:
 \begin{equation}
\label{eq:unbalanced-wasserstein}
\uot(\bx, \by) = \min_{\bP \in {\bbR_+}^{p\times p}} \, \varepsilon \kl(\bP| e^{- \frac{\bC}{\varepsilon}}) + \gamma \kl(\bP\mathds 1 | \bx) + \gamma \kl(\bP^\top \mathds 1 | \by) \enspace.
\end{equation}
The following proposition shows how to compute UOT using a generalized version of Sinkhorn's algorithm.
\begin{proposition}
\label{p:dualw}
Let $\bx, \by \in \bbR^p_+$. The unbalanced Wasserstein distance is obtained from the dual problem:
\begin{equation}
    \label{eq:dualw}
  \uot(\bx, \by) = \max_{u, v \in \bbR^p} - \gamma \langle \bx, e^{-\frac{u}{\gamma}} - 1\rangle - \gamma \langle \by, e^{-\frac{v}{\gamma}} - 1\rangle -  \varepsilon \langle e^{\frac{u \oplus v}{\varepsilon}} - 1, e^{-\frac{\bC}{\varepsilon}}\rangle \enspace .
\end{equation}
Moreover, with the change of variables: 
$\omega = \frac{\gamma}{\gamma + \varepsilon}$, $\bK=e^{-\frac{\bC}{\varepsilon}}, \ba = e^{\frac{u}{\varepsilon}}, \bb = e^{\frac{v}{\varepsilon}}$, the optimal dual points are the solutions of the fixed point problem:
\begin{equation}
    \label{eq:fixedpoint}
      \ba = \left(\frac{\bx}{\bK\bb}\right)^{\omega} \quad , \quad
    \bb = \left(\frac{\by}{\bK^{\top}\ba}\right)^{\omega}
\end{equation}

and the optimal transport plan is given by: \begin{equation}
    \label{eq:primaldual}
(\bP_{ij}) = (\ba_i\bK_{ij}\bb_j). 
\end{equation}

\end{proposition}

\proof
Since the conjugate of the linear operator $G:\bP \mapsto (\bP\mathds1, \bP^{\top}\mathds 1)$ is given by $G^\star: (u, v) \mapsto u \oplus v$, the Fenchel duality theorem leads to \eqref{eq:unbalanced-wasserstein}. The dual loss function is concave and goes to $-\infty$ when $\|u, v\| \to +\infty$, canceling its gradient yields \eqref{eq:fixedpoint}.
Finally, since the primal problem is convex, strong duality holds and the primal-dual relationship gives \eqref{eq:primaldual}. See \cite{chizat17} for a detailed proof.
\hfill $\square$

 Solving the fixed point problem \eqref{eq:fixedpoint} is equivalent to alternate maximization of the dual function \eqref{eq:dualw}. Starting from two vectors $\ba, \bb$ set to $\mathds 1$, the algorithm iterates through the scaling operations \eqref{eq:fixedpoint}. This is a generalization of the Sinkhorn algorithm which corresponds to  $\omega = 1$ or $\gamma = +\infty$.
\begin{corollary}
	\label{cor:symmetricw}
	Let $\bx \in \bbR^p_+$. The associated optimal dual scalings $\ba$, $\bb$ to computing $\uot(\bx, \bx)$ are given by the solution of the fixed point problem:
	$ \bb = \ba = \left(\frac{\by}{K\ba}\right)^{\phi} $
\end{corollary}
\proof The symmetry of the dual problem \eqref{eq:dualw} with $\bx = \by$ implies immediately that $\ba = \bb$. Proposition \ref{p:dualw} gives the fixed point equation.

\paragraph{Debiased UOT}
Due to regularization, UOT is known to suffer from an entropy bias causing UOT to fail to identify identical distributions $\uot(\bx, \bx) \neq 0$. Inspired by the balanced OT case \cite{feydy19}, we define the debiased UOT cost:
\begin{equation}
\label{eq:sdivergence}
\widetilde{\uot}(\bx, \by) = \uot(\bx, \by) - \frac{1}{2}\left(\uot(\bx, \bx) + \uot(\by, \by)\right)
\end{equation}
$\widetilde{\uot}$ is coercive with respect to each of its arguments, moreover, if $\bK \eqdef e^{-\frac{\bC}{\varepsilon}}$ is positive semi-definite, then $S$ is non-negative~\citep{janati20-aistats}. Therefore, $\sdtw$ is well defined with $\widetilde{\uot}$ as a cost function, a loss function we named STA: \emph{Spatio-Temporal Alignment}.
\begin{definition}[STA]
	\label{def:wdtw}
	We define the STA loss as:
	\begin{equation}
	\label{eq:wdtw}
	\sta(\bx, \by) = \sdtw(\bx, \by; \widetilde{\uot})
	\end{equation}
\end{definition}

\begin{remark}
	It is important to note that all the properties shown for $\widetilde{\uot}$ (coercivity, non-negativity) hold only if the reference measure used to define the entropy regularization is the uniform distribution over a fixed universal support $\cA$ for all measures i.e the penalty applied to the transportation plan writes $E(\bP) \eqdef \kl(\bP | \mathds 1_p / p)$. For instance, when considering the more general formulation with the product measure as a reference ($\kl( \bP |\bx\otimes\by)$), \citet{sejourne19} showed that non-negativity and convexity hold but with an additional quadratic term between the masses of the distributions $\frac{\varepsilon}{2}(\bx^\top\mathds1 - \by^\top\mathds1)^2$ in the definition of $\widetilde{\uot}$.
\end{remark}
\subsection{Computing barycenters with $\widetilde{\uot}$}
To use algorithm \ref{a:barycenter}, it is required to compute barycenters with the inner $\sdtw$ cost function $\widetilde{\uot}$ as dissimilarity. In the following, we show that this barycenter can be estimated using a modified Sinkhorn algorithm.
Let $\bx_1, \dots, \bx_K \in \bbR^{p}_+$ and $w_1, \dots, w_K$ a sequence of positive weights adding to 1.
$\widetilde{\uot}$ is non-negative and coercive, thus its barycenter problem is well defined:
\begin{equation}
\label{eq:delta-bar}
\min_{\bx\in \bbR^p_+} \mathcal{J}(\bx) \eqdef \min_{\bx \in \bbR^p_+}\sum_{k=1}^K w_k \widetilde{\uot}(\bx_k, \bx)
\end{equation}
$\uot$ is differentiable, and its gradient is given by  $\gamma (1 - \ba^{-\frac{\varepsilon}{\gamma}}, 1 - \bb^{-\frac{\varepsilon}{\gamma}})$ where $(\ba, \bb)$ is the solution of the fixed equation \eqref{eq:fixedpoint} \cite{feydy17}. Thus, using the chain rule, $\mathcal{J}$ is also differentiable and its gradient is given by:
\begin{equation}
\label{eq:grad-J}
\nabla \mathcal{J}(\bx) = \gamma (\bc^{-\frac{\varepsilon}{\gamma}} - \sum_{k=1}^K w_k \bb_k^{-\frac{\varepsilon}{\gamma}})
\end{equation}
where, using the notations of Proposition \ref{p:dualw}, $\bc, \bb_1, \dots,  \bb_K, \ba_1, \dots, \ba_K \in \bbR^p_+$ verify the fixed point equations:
\begin{equation}
\label{eq:fixedpoint-grad}
\ba_k = \left(\frac{\bx_k}{\bK\bb_k}\right)^{\omega} \quad , \quad
\bb_k = \left(\frac{\bx}{\bK^{\top}\ba_k}\right)^{\omega} \quad, \quad 
\bc = \left(\frac{\bx}{\bK\bc}\right)^{\omega}
\end{equation}
Without studying the convexity of $\mathcal{J}$, we can show that any stationary point of $\mathcal{J}$ is actually a global minimum. Thus, it is sufficient to solve $\nabla \cJ (\bx) = 0$ to compute the $\widetilde{\uot}-$barycenter. The following lemma plays a major role in proving this statement.
\begin{lemma}[Suboptimality]
\label{lem:suboptimality}

Let $\bx, \by \in \bbR^p_+$. Let $\ba, \bb \in \bbR^p_+$ be the optimal dual variables associated with $\uot(\bx, \by)$ i.e
the solutions of the optimality conditions $\ba = \left(\frac{\bx}{\bK\bb}\right)^{\omega}$ and $
 \bb = \left(\frac{\by}{\bK^{\top}\ba}\right)^{\omega}$. Then for any $\bbf, \bg \in \bbR^p_+$: 
 \begin{equation}
 \label{eq:suboptimality}
 \gamma \langle \bx, \bbf^{-\frac{\varepsilon}{\gamma}} \rangle +\gamma \langle \by, \bg^{-\frac{\varepsilon}{\gamma}} \rangle + \varepsilon\langle \bbf, \bK\bg\rangle \geq (\varepsilon + 2\gamma)\langle \ba, \bK \bb \rangle
 \end{equation}
\end{lemma}
\proof
Using the same change of variable $\bbf = e^{\frac{u}{\varepsilon}}, \bg = e^{\frac{v}{\varepsilon}}$, the dual problem of Proposition \ref{p:dualw} can be written:
\begin{align*}
\uot(\bx, \by) &= \max_{\bbf, \bg \in \bbR_+^p} - \gamma \langle \bx, \bbf^{-\frac{\varepsilon}{\gamma}} - 1\rangle - \gamma \langle \by, \bg^{-\frac{\varepsilon}{\gamma}}  - 1\rangle -  \varepsilon \langle \bbf \otimes \bg - 1, \bK\rangle \\ 
					  &=\max_{\bbf, \bg \in \bbR_+^p} - \gamma \langle \bx, \bbf^{-\frac{\varepsilon}{\gamma}} - 1\rangle - \gamma \langle \by, \bg^{-\frac{\varepsilon}{\gamma}}  - 1\rangle -  \varepsilon \langle \bbf \otimes \bg - 1, \bK\rangle \\
					    &=\max_{\bbf, \bg \in \bbR_+^p} - \gamma \langle \bx, \bbf^{-\frac{\varepsilon}{\gamma}} \rangle - \gamma \langle \by, \bg^{-\frac{\varepsilon}{\gamma}}  \rangle -  \varepsilon \langle \bbf, \bK\bg \rangle - \varepsilon\|\bK\|_1  + \gamma (\|\bx\|_1 + \|\by\|_1) \\
\end{align*}
Since $\ba, \bb$ are the solution of the dual problem above, at optimality it holds:
\begin{align*}
\langle \bx, \ba^{-\frac{\varepsilon}{\gamma}} \rangle = \langle\ba^{\frac{1}{\omega}} \bK\bb, \ba^{-\frac{\varepsilon}{\gamma}} \rangle  = \langle \ba, \bK\bb \rangle 
\end{align*}
Similarly:
\begin{align*}
\langle \by, \bb^{-\frac{\varepsilon}{\gamma}} \rangle = \langle\bb^{\frac{1}{\omega} }\bK^\top\ba, \bb^{-\frac{\varepsilon}{\gamma}} \rangle  = \langle \bb, \bK^\top\ba \rangle  = \langle \ba, \bK\bb \rangle 
\end{align*}
Thus:
\begin{equation*}
\uot(\bx, \by) = -  (\varepsilon + 2\gamma) \langle \ba, \bK\bb \rangle - \varepsilon\|\bK\|_1  + \gamma (\|\bx\|_1 + \|\by\|_1)
\end{equation*}
By the definition of the max operator, it holds for any $\bbf, \bg \in \bbR^p_+$:
\begin{equation*}
 \gamma \langle \bx, \bbf^{-\frac{\varepsilon}{\gamma}} \rangle +\gamma \langle \by, \bg^{-\frac{\varepsilon}{\gamma}} \rangle + \varepsilon\langle \bbf, \bK\bg\rangle \geq (\varepsilon + 2\gamma)\langle \ba, \bK \bb \rangle
\end{equation*}
\hfill $\square$

Since $\cJ$ is coercive, it has at least one global minimum. The following proposition shows that this minimum is unique.
\begin{proposition}
	\label{prop:minimum}
	Let $\bar{\bx} \in \bbR_+^p$ such that $\nabla \cJ(\bar{\bx}) = \boldsymbol{0}$. Then for any $\by \in \bbR^p_+$ it holds: 
	$$ \cJ(\by) \geq \cJ(\bar{\bx})$$. 
\end{proposition}
\proof
Let $\bd_1, \dots, \bd_K$ the symmetric dual variables used to compute $\widetilde{\uot}(\bx_k, \bx_k)$ for $k=1..K$ i.e
the solutions of $\bd_k = \left( \frac{\bx_i}{\bK\bd_k}\right)^\omega$. 
 Let $\by \in \bbR^p_+$ and its associated dual variables $\bc', \ba'_1, \dots, \ba'_K, \bb'_1, \dots, \bb'_K$  used to compute $\widetilde{\uot}(\by, \by), \widetilde{\uot}(\bx_1, \by), \dots, \widetilde{\uot}(\bx_K, \by)$. Therefore,  it holds:
\begin{align*}
\cJ(\by) &= (\varepsilon + 2\gamma) \sum_{k=1}^K w_k \left( \frac{1}{2}(\langle  \bc', \bK\bc'\rangle + \langle  \bd_k, \bK\bd_k\rangle)  - \langle \ba'_k, \bK\bb'_k\rangle\right) 
\end{align*}
Let $\bc, \ba_1, \dots, \ba_K, \bb_1, \dots, \bb_K$ denote the dual variables verifying \eqref{eq:fixedpoint-grad} for $\bx = \bar{\bx}$. Moreover, since $\nabla \cJ(\bar{\bx}) = \boldsymbol{0}$, it holds: $\bc^{-\frac{\varepsilon}{\gamma}} = \sum_{k=1}^K w_k \bb_k^{-\frac{\varepsilon}{\gamma}} $, therefore:
\begin{equation}
\label{eq:grad-bar-0}
\sum_{k=1}^K w_k \langle \ba_k, \bK\bb_k\rangle = \sum_{k=1}^K w_k \langle \bar{\bx}, {\bb_k}^{-\frac{\varepsilon}{\gamma}} \rangle = \langle \bar{\bx}, {\bc}^{-\frac{\varepsilon}{\gamma}} \rangle = \langle \bc, \bK\bc\rangle 
\end{equation}
Thus, evaluate $\cJ$ at $\bar{\bx}$ leads to:
\begin{align*}
\cJ(\bar{\bx}) &= (\varepsilon + 2\gamma) \sum_{k=1}^K w_k \left( \frac{1}{2}(\langle  \bc, \bK\bc\rangle + \langle  \bd_k, \bK\bd_k\rangle)  - \langle \ba_k, \bK\bb_k\rangle\right) \\
&=\frac{1}{2} (\varepsilon + 2\gamma) \left(\sum_{k=1}^K w_k  \langle  \bd_k, \bK\bd_k\rangle - \langle \bc, \bK\bc\rangle\right) 
\end{align*}
Thus, the statement we wish to prove is equivalent to:
\begin{align}
\label{eq:ineq-eq}
\cJ(\by) \geq \cJ(\bar{\bx}) \Leftrightarrow \frac{1}{2}(\varepsilon + 2\gamma)  \left(\langle  \bc', \bK\bc'\rangle + \langle  \bc, \bK\bc\rangle\right)  \geq (\varepsilon + 2\gamma)  \sum_{k=1}^K w_k \langle \ba'_k, \bK\bb'_k\rangle
\end{align}
For each element of the sum in the right side above, let's derive un upper bound using Lemma~\ref{lem:suboptimality}. 
Consider the sub-optimal dual variables $(\bbf_k, \bg_k) = (\ba_k, \bb_k \odot \frac{\bc'}{\bc})$. It holds:
\begin{align}
\label{eq:lemma-applied}
 \gamma \langle \bx_k, {\ba_k}^{-\frac{\varepsilon}{\gamma}} \rangle +\gamma \langle \by, (\bb_k\odot \frac{\bc'}{\bc})^{-\frac{\varepsilon}{\gamma}} \rangle + \varepsilon\langle \bb_k \odot \frac{\bc'}{\bc} , \bK^\top \ba_k\rangle \geq (\varepsilon + 2\gamma)  \langle \ba'_k , \bK\bb'_k\rangle 
\end{align}
Applying the weighted sum and using the optimality conditions along with $\cJ(\bar{\bx}) = 0$, the elements in the left side can be further simplified as:
\begin{align*}
\sum_{k=1}^K w_k \langle \bx_k, {\ba_k}^{-\frac{\varepsilon}{\gamma}} \rangle &= \sum_{k=1}^K w_k \langle \bar{\bx}, {\bb_k}^{-\frac{\varepsilon}{\gamma}} \rangle = \sum_{k=1}^K w_k \langle \bar{\bx}, {\bc}^{-\frac{\varepsilon}{\gamma}} \rangle = \langle \bc, \bK\bc \rangle \\
\sum_{k=1}^K w_k \langle \by, (\bb_k \odot \frac{\bc'}{\bc})^{-\frac{\varepsilon}{\gamma}} \rangle &=  \langle \by  \odot (\frac{\bc'}{\bc})^{-\frac{\varepsilon}{\gamma}}  , \sum_{k=1}^K w_k \bb_k^{-\frac{\varepsilon}{\gamma}} \rangle = 
\langle \by  \odot (\frac{\bc'}{\bc})^{-\frac{\varepsilon}{\gamma}} ,  {\bc}^{-\frac{\varepsilon}{\gamma}} \rangle  = \langle \by, {\bc'}^{-\frac{\varepsilon}{\gamma}} \rangle = \langle \bc', \bK\bc'\rangle\\
\sum_{k=1}^K w_k \langle \bb_k \odot \frac{\bc'}{\bc} , \bK^\top \ba_k\rangle &=  \langle  \frac{\bc'}{\bc} \odot \bar{\bx} , \sum_{k=1}^K w_k {\bb_k}^{-\frac{\varepsilon}{\gamma}}\rangle  = \langle \bc' \odot \bar{\bx} , \bc^{-\frac{\gamma + \varepsilon}{\gamma}}\rangle = \langle \bc' , \bK\bc\rangle 
\end{align*}
Therefore, summing over equation \eqref{eq:lemma-applied}:
\begin{align*}
\gamma \langle \bc, \bK\bc \rangle  + \gamma \langle \bc', \bK\bc'\rangle + \varepsilon \langle \bc' , \bK\bc\rangle \geq (\varepsilon + 2\gamma)  \sum_{k=1}^K w_k \langle \ba'_k , \bK\bb'_k\rangle 
\end{align*}
On another side, since $\bK$ is positive semi-definite, it holds:
\begin{align*}
\label{eq:positivity}
\langle \bc - \bc', \bK(\bc - \bc')\rangle \geq 0 \Rightarrow  \frac{1}{2}(\langle \bc, \bK\bc \rangle  + \langle \bc', \bK\bc'\rangle) \geq  \langle \bc', \bK\bc\rangle
\end{align*}
Combining the last two inequalities leads to \eqref{eq:ineq-eq} ending the proof.
\hfill $\square$.

\begin{figure}
	\includegraphics[trim={3cm 1cm 2.5cm 0.cm}, clip, width=0.9\linewidth]{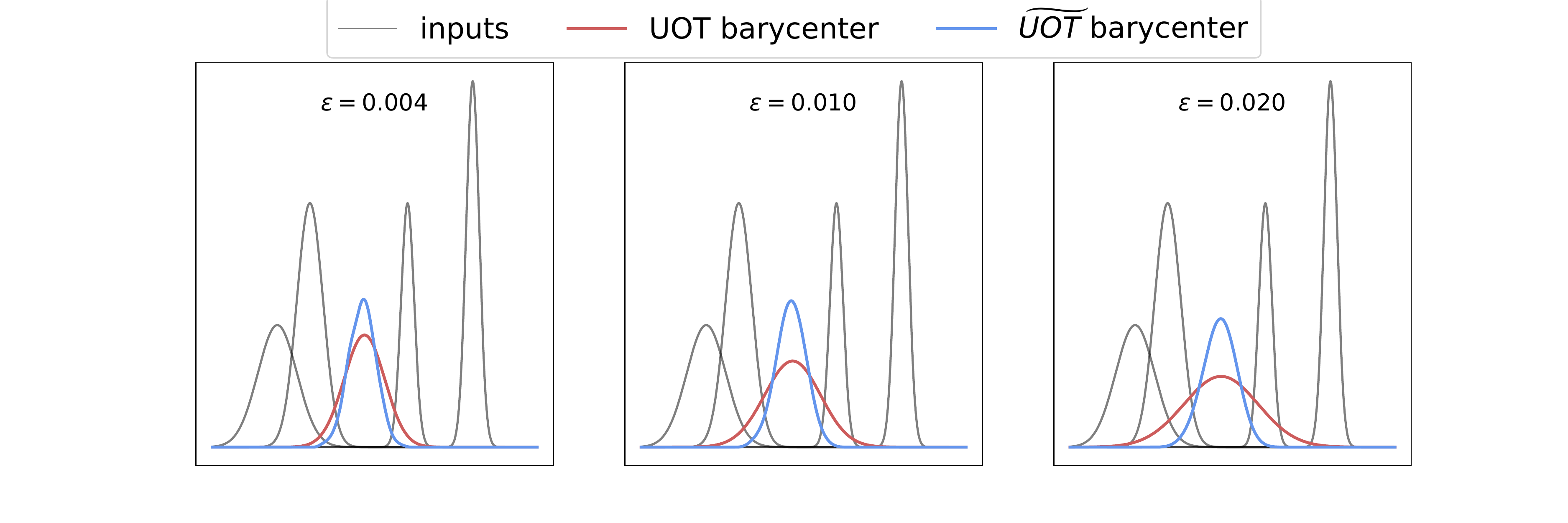}
	\caption{Barycenters of 4 Gaussian distributions with different means, variances and total masses. The debiased barycenter $\widetilde{\uot}$ is less sensitive to entropy regularization ($\varepsilon$) than the original
		unbalanced barycenter UOT. \label{f:gaussians}}
\end{figure}

To solve the barycenter problem \eqref{eq:delta-bar}, it is sufficient to solve the fixed point system:
\begin{equation}
\label{eq:fixedpoint-delta}
\ba_k = \left(\frac{\bx_k}{\bK\bb_k}\right)^{\omega} \quad , \quad
\bb_k = \left(\frac{\bar{\bx}}{\bK^{\top}\ba_k}\right)^{\omega} \quad, \quad 
\bc = \left(\frac{\bar{\bx}}{\bK\bc}\right)^{\omega}\quad, \quad
\sum_{k=1}^K w_k {\bb_k}^{-\frac{\varepsilon}{\gamma}} = \bc^{-\frac{\varepsilon}{\gamma}}
\end{equation}
which --  combining the last 3 equations -- is equivalent to:
\begin{equation}
\label{eq:fixedpoint-delta2}
\ba_k = \left(\frac{\bx_k}{\bK\bb_k}\right)^{\omega} \,, \quad
\bb_k = \left(\frac{\bar{\bx}}{\bK^{\top}\ba_k}\right)^{\omega} \, , \quad 
\bc = \left(\frac{\bar{\bx}}{\bK\bc}\right)^{\omega}\, , \quad
\bar{\bx} = \bc^{\frac{1}{\omega}} \left(\sum_{k=1}^K w_k (\bK^\top \ba_k)^{1 - \omega}\right)^{\frac{1}{1-\omega}}
\end{equation}
These equations are very similar to the barycentric Sinkhorn algorithm of \citet{chizat17}. Indeed, disregarding the symmetric equation in $\bc$ and setting $\bc = \mathds 1_p$ in the update of $\bar{\bx}$, we recover Sinkhorn's iterations for the UOT barycenter. These updates lead to Algorithm \ref{alg:sinkhorn}. While the theoretical analysis of its convergence is left for future work, we empirically observe that it converges regardless of the initialization of the dual variables. More importantly, it leads sharper barycenter than the (biased) UOT barycenters for almost no additional computational cost. Figure \ref{f:gaussians} shows the example of Gaussians with different means, variances and masses for 3 values of $\varepsilon$. 
 \begin{algorithm}
	\caption{Debiased unbalanced $\widetilde{\uot}$ barycenter.\label{alg:sinkhorn}}
	\begin{algorithmic}
		\STATE {\bfseries Input: $\bx_1, \dots, \bx_K \in \bbR^p_+$,  parameters $\varepsilon, \gamma > 0$, $\bK \eqdef  e^{-\frac{\bM}{\varepsilon}}$}
		\STATE {\bfseries Output: $\bar{\bx},$ the $\widetilde{\uot}$ barycenter of $(\bx_1, \dots, \bx_K)$}
		\STATE {\bfseries Initialize $\bc = \bb_1 = \dots = \bb_K = \mathds 1_p$, set $\omega = \frac{\gamma}{\gamma + \varepsilon}$}
		\WHILE{Not converged}
		\FOR{$k=1$ {\bfseries to} $K$}
		\STATE{ $\ba_k = \left(\frac{\bx_k}{\bK\bb_k}\right)^{\omega}$}
		\ENDFOR
		\STATE{$\bar{\bx} = \bc^{\frac{1}{\omega}} \left(\sum_{k=1}^K w_k {(\bK^\top\ba_k)}^{1 -\omega}\right)^{\frac{1}{1-\omega}}$}
		\FOR{$k=1$ {\bfseries to} $K$}
		\STATE{ $\bb_k = \left(\frac{\bar{\bx}}{\bK^\top\ba_k}\right)^{\omega}$}
		\ENDFOR
		\STATE{$\bc =  \left( \frac{\bar{\bx}}{\bK\bc}\right)^{\omega}  $}
		\ENDWHILE
	\end{algorithmic}
\end{algorithm}

\begin{remark}
	The proposition \ref{prop:minimum} may seem indicate that $\mathcal{J}$ has a positive curvature. However, it is easy to show that $\widetilde{\uot}$ is not convex in dimension 1. Indeed, taking $p=1$ leads to $\bK = 1$ and the Sinkhorn equations can be solved in closed form. We obtain for $x, y \in \bbR_+$: 
	$$\widetilde{\uot}(x, y) = (\varepsilon+ 2\gamma ) \left(\frac{x^{\frac{2\omega}{\omega + 1}} + y^{\frac{2\omega}{\omega + 1}}}{2} - (xy)^{\frac{\omega}{\omega + 1}} \right)$$
	Since $\omega < 1$, in dimension 1, $x \mapsto \widetilde{\uot}(x, y)$ is strictly concave provided $x$ is large enough.
\end{remark}

\paragraph{Complexity analysis}
As shown by Algorithm \ref{a:dynamicprogram}, soft-DTW is quadratic in time. Computing the $\widetilde{\uot}$ pairwise matrix is quadratic in $p$. Moreover, when the time series are defined on regular grids such as images, one could benefit from spatial Kernel separability as introduced in \cite{solomon15}. This trick allows to reduce the complexity of Algorithm \ref{alg:sinkhorn} on 2D data from $O(p^2)$ to $O(p^{\frac{3}{2}})$. Moreover, to leverage fast matrix products on GPUs, computing the $\widetilde{\uot}$ barycenter and the pairwise $\widetilde{\uot}$ distance matrices  can be done in parallel so that the iterations within both k-loops in Algorithm \ref{alg:sinkhorn} are run simultaneously.


\section{Experiments}
\label{s:experiments}
We illustrate the effectiveness of the STA barycenter in two experiments with real data.  Python code can be found in \href{https://github.com/hichamjanati/spatio-temporal-alignements}{https://github.com/hichamjanati/spatio-temporal-alignements}.

\paragraph{Optimal transport hyperparameters}
The debiased divergence $\widetilde{\uot}$ is defined by the same hyperparameters of $\uot$: $\varepsilon$ and $\gamma$. Given that some of the entropy bias is removed with $\widetilde{\uot}$, the obtained barycenter is less sensitive to $\varepsilon$ than the original UOT barycenter (Fig \ref{f:gaussians}). However, setting $\varepsilon$ too large slows down the convergence of the symmetric potential $\bc$. Here we set $\varepsilon = 1. / p$. The marginals parameter $\gamma$ must be large enough to guarantee transportation of mass. When $\gamma \to 0$, the optimal transport plan $\bP^\star \to \bK$. Large $\gamma$ however slows down the convergence of Sinkhorn's algorithm, especially if the input histograms have significantly different total masses that are concentrated far from each other. We set $\gamma$ at the largest value guaranteeing a minimal transported mass using the heuristic proposed in \cite{janati19}.

\subsection{Averaging of brain imaging data}
\begin{figure}[t]
	\centering
\includegraphics[width=\linewidth]{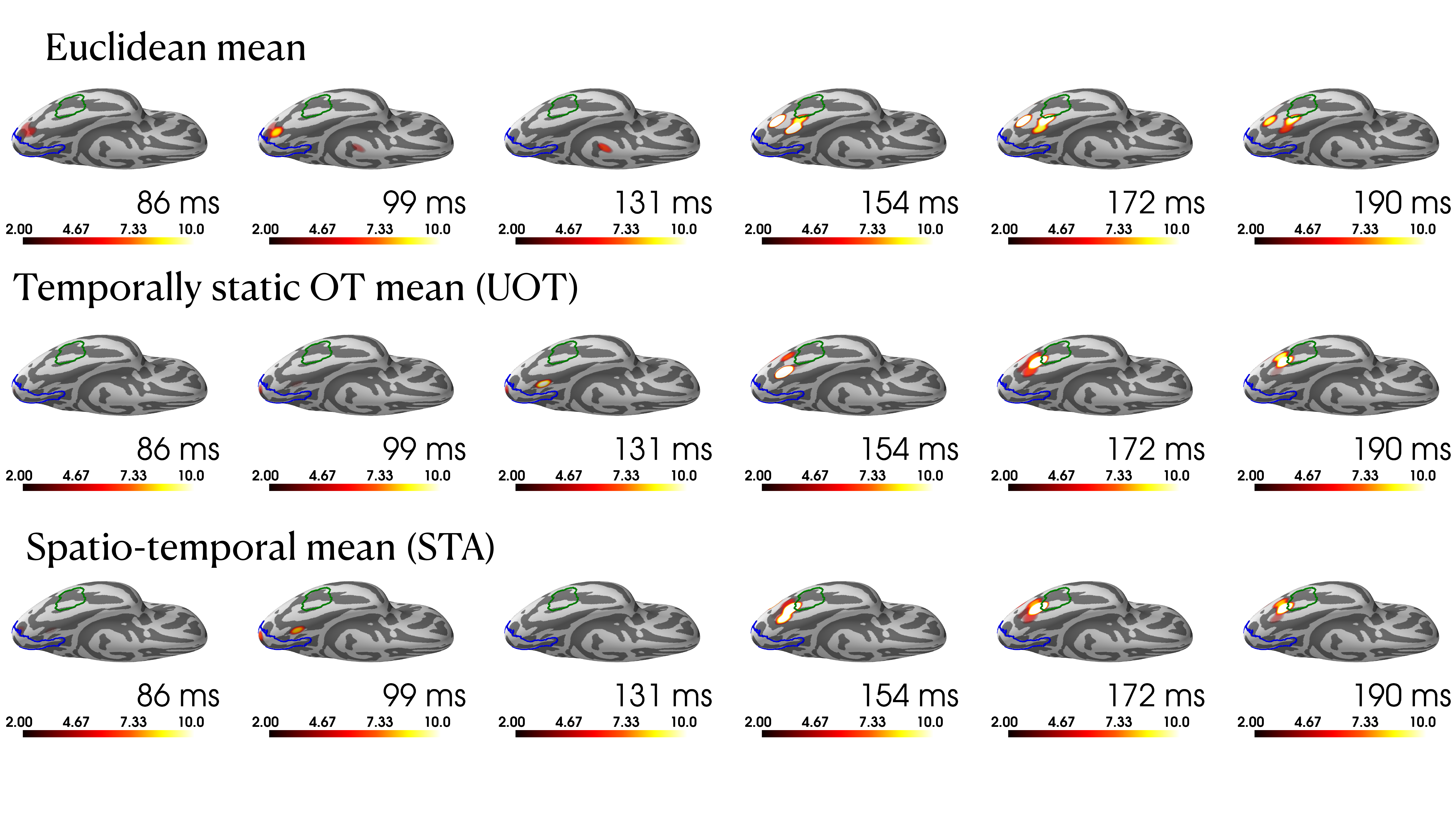}
\caption{Barycenters of the spatio-temporal neural activity of 6 subjects taken from the DS117 dataset. The STA barycenter shows a more focused activation around the Fusiform Form Face Area (green) than the other methods. Unlike the OT barycenter, STA shows a more plausible time occurrence of the first evoked response around 100ms.  \label{f:brains}}
\end{figure}

\begin{figure}
\centering
\includegraphics[width=0.8\linewidth]{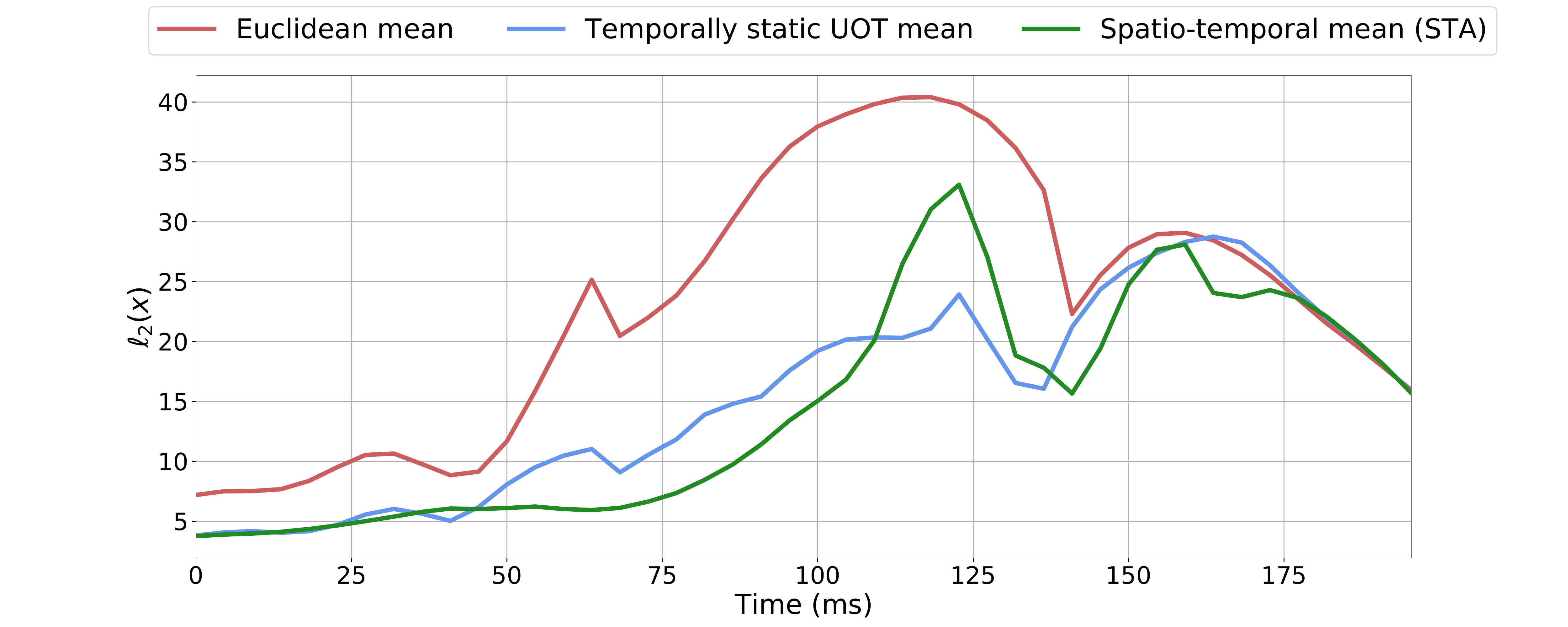}
\caption{$\ell_2$ norm (across space) of the barycenters shown in Figure \ref{f:brains} showing a clear temporal sensitivity of STA as it identifies the two expected evoked responses to visual facial stimulus. \label{f:time-norms}}
\end{figure}

Studying the function of the various regions of the Human brain is one of the primary goals of neuroimaging research. These studies usually involve a group of healthy individuals (subjects) or patients who perform a series of tasks while having their neural activity recorded from which active regions of the brain are localized. However, drawing conclusions at a population level requires an aggregation function that combines the individual active sources of each subject. While averaging may seem like a straightforward and simple solution, it does not take into account the anatomical differences across subjects which lead to spatially blurred means. Moreover, the brain responses of the different subjects are never synced in time, specially when working with Electro-encephalogrphy (EEG) or Magneto-encephalography (MEG) data which have a high temporal resolution of the order of 1 millisecond. We use public the EEG/MEG dataset DS117~\citep{ds117} and compute the spatio-temporal source configuration of 6 subjects who were shown images of Human faces using MNE-Python~\citep{mne-python}. Here the support of our measures $\cA$ is taken to be the set of 642 vertices that define the cortical mesh of the brain. The OT ground metric $\bC$ is defined as the quadratic length of the shortest path on the triangulated mesh.  We compute 3 different averages: a Euclidean mean, a $\widetilde{\uot}$ barycenter (independently across time) and a spatio-temporal STA barycenter
with $k_{\max} = 20$. As shown in Figure \ref{f:brains}, the first burst in the neural response is a visually evoked potential (known as P1) that arises around 100ms after the stimulus~\citep{slotnick99} in the primary visual cortex (blue). Then, at around 170ms, an evoked response that is specific to the display of faces occurs in a small region known as the Fusiform Face Area (Green) (FFA)~\citep{bentin96, kanwisher97}. The delimited regions of interest were selected using the meta-analysis tool Neurosynth~\citep{neurosynth}.
To further assess the temporal sensitivity of STA, we display in Figure \ref{f:time-norms} the $\ell_2$ norm across space of the 3 barycenters of Figure \ref{f:brains}. The two evoked responses are more pronounced when using the STA barycenter.

\subsection{Forecasting the motion of handwritten letters}

\paragraph{Dataset}
We evaluate the performance of STA in a prediction task using a publicly available dataset of handwritten letters where the position of a pen are tracked in time \cite{williams}. We subsample the data both spatially and temporally so as to keep 13 time points of (30$\times$30) images for each time series. Each image can thus be seen as a screenshot at a certain time during the writing motion. To make the task a bit more challenging, we randomly shift each time series spatially (resp. temporally) by 0 to 10 pixels in each direction (by keeping 5 to 13 time points evenly selected). The dataset is composed of 20 samples of each one of the letters (``a'', ``b'', ``c'', ``v''), thus the full shape of the dataset is (100, 13, 30, 30). 

\begin{figure}[t]
	\centering
	\includegraphics[width=0.8\linewidth]{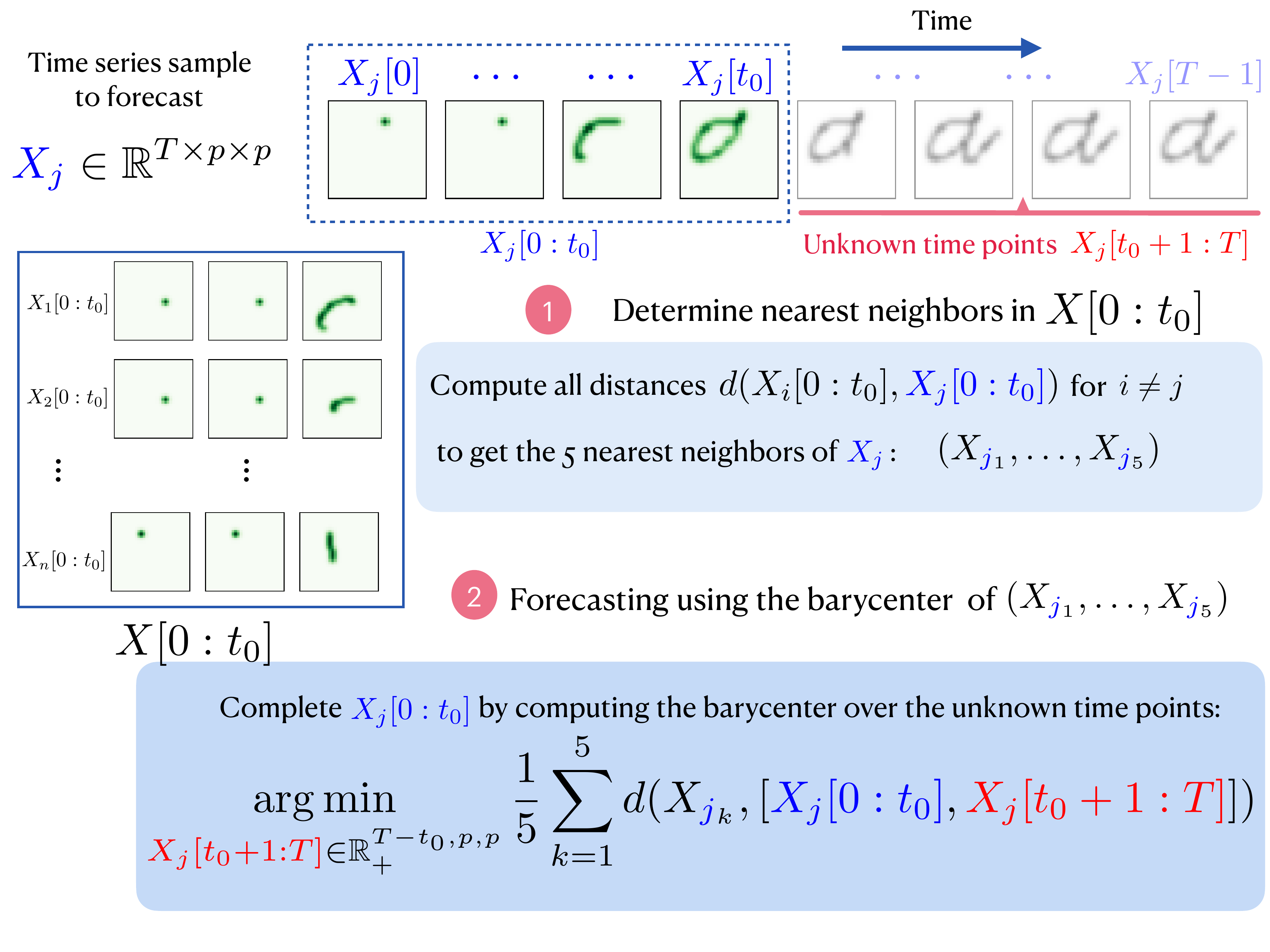}
	\caption{Sketch explaining the forecasting pipeline used with the handwritten letters experiment. \label{f:pipeline}}
\end{figure}
\begin{figure}
	\centering
\includegraphics[width=0.8\linewidth]{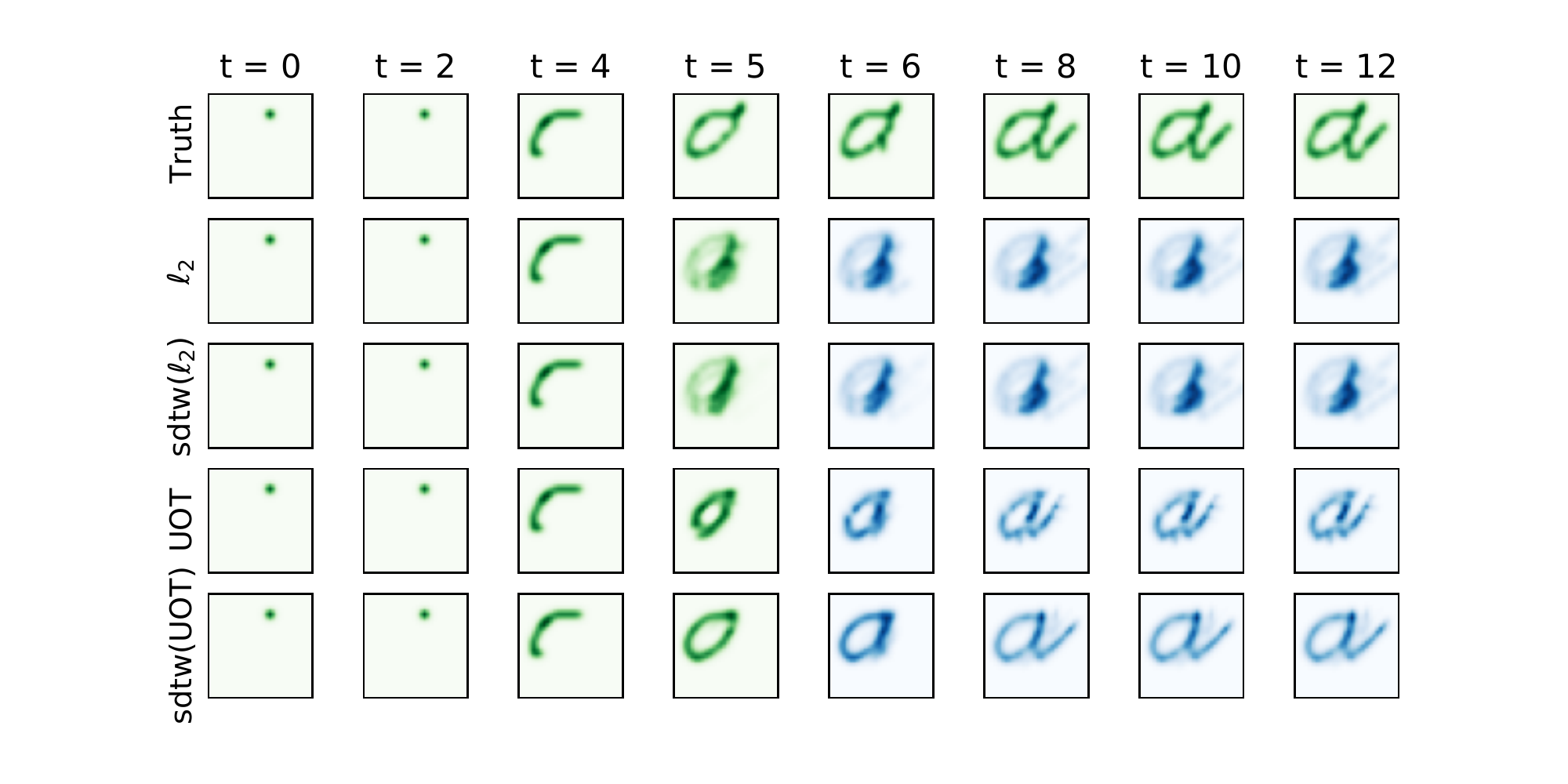}
\caption{Forecasts of a handwritten letter time series. The green time points are fixed and considered known for all models. Blue observations are predicted. As expected $\ell_2$ based methods fail to identify neighbors in the same class. \label{f:predictions}}
\end{figure}
\begin{figure}
	\centering
\includegraphics[width=0.8\linewidth]{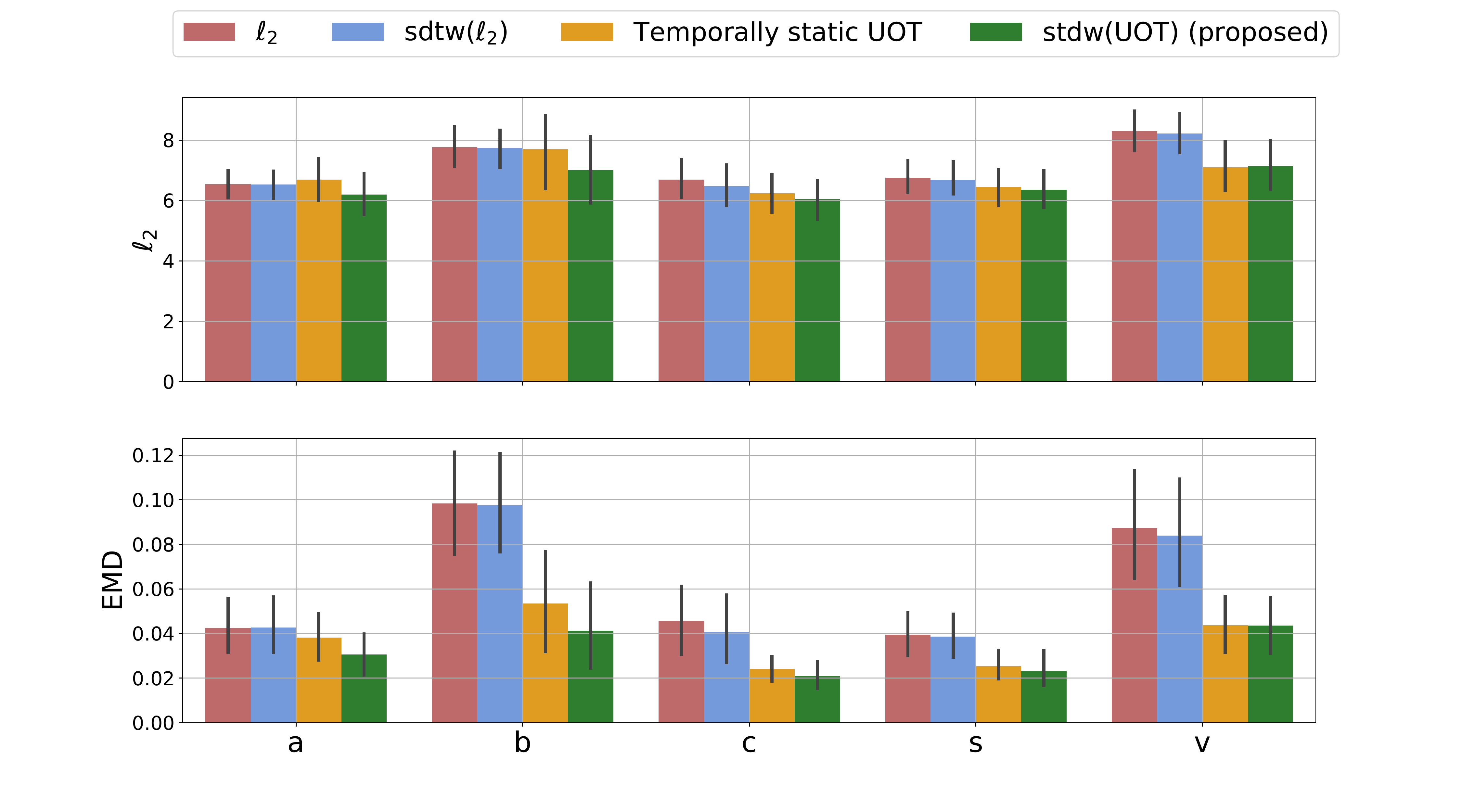}
\vspace{-0.3cm}
\caption{Mean prediction scores computed on the unknown halves of the time series for each letter. The EMD score is computed between each true image and its prediction after normalization to the probability simplex. The ground metric is the Euclidean distance between pixel locations, normalized so that EMD is within [0-1]. \label{f:scores}}
\end{figure}

\paragraph{Forecasting}
We propose to use barycenters as a forecasting method. . For each time series $\bx$ in the dataset, knowing only the first $t_0 < 13$ time points, we would like to predict the rest. First, based only on the observed $t_0 = 5$ time points, we select the closest 5 neighbors of $\bx$ in the data based on some loss function $d$. We denote these nearest neighbors $\bx'_1, \dots, \bx'_5$. Next, we predict the future of $\bx$ by computing the $d$-barycenter of $(\bx'_k)_{k=1..5}$ while keeping the first $t_0$ observations of $\bx$ fixed. The full pipeline is illustrated in Figure \ref{f:pipeline}. The predictions obtained for the example shown in Figure \ref{f:pipeline} are illustrated in Figure \ref{f:predictions}. While $\ell_2$ based method clearly fail to identify neighbors in the same class (``a''), OT based methods do not. Moreover, thanks to temporal variability, STA provides a more accurate prediction of the remaining time points than OT alone.

Figure \ref{f:scores} shows a more quantitative comparison where we evaluate the accuracy of the predictions for all samples in the dataset with the $\ell_2$ and the EMD (Earth mover distance) metrics averaged across the 8 predicted time points. To compute the EMD scores, we normalize all images so that their values add up to 1 and define EMD with the Euclidean quadratic cost between pixel coordinates. On both metrics and for all letters, STA outperforms the other loss functions. 
%


\section{Conclusion}
Spatio-temporal data can differ in amplitude and in spatio-temporal structure. To average such data we proposed a combination of Soft-DTW and optimal transport. Our main contributions are twofold. First, we derive a fast alternating optimization schedule that reduces the spatio-temporal averaging to a sequence of temporally weighted spatial averages. Second, combined with entropic unbalanced OT as a spatial mean, we show that the debiased version can also be used to compute barycenters using a generalized version of Sinkhorn's algorithm. The performance of our experiments on simulations and real data confirm our findings and show that our method can identify meaningful spatio-temporal barycenters.

\subsection*{Acknowledgements}
MC and HJ acknowledge the support of a chaire
d’excellence de l’IDEX Paris Saclay. AG and HJ were supported by the European Research Council Starting Grant SLAB ERC-StG-676943.

\bibliography{references}

\begin{thebibliography}{30}
\providecommand{\natexlab}[1]{#1}
\providecommand{\url}[1]{\texttt{#1}}
\expandafter\ifx\csname urlstyle\endcsname\relax
  \providecommand{\doi}[1]{doi: #1}\else
  \providecommand{\doi}{doi: \begingroup \urlstyle{rm}\Url}\fi

\bibitem[Bentin et~al.(1996)Bentin, Allison, Puce, Perez, and
  McCarthy]{bentin96}
Shlomo Bentin, Truett Allison, Aina Puce, Erik Perez, and Gregory McCarthy.
\newblock Electrophysiological studies of face perception in humans.
\newblock \emph{Journal of Cognitive Neuroscience}, 8\penalty0 (6):\penalty0
  551--565, 1996.

\bibitem[Chen and Qi(2003)]{chen03}
Chao-Ping Chen and Feng Qi.
\newblock The best bounds of harmonic sequence.
\newblock \emph{arXiv preprint math/0306233}, 2003.

\bibitem[Chizat et~al.(2017)Chizat, Peyr{\'e}, Schmitzer, and
  Vialard]{chizat17}
L.~Chizat, G.~Peyr{\'e}, B.~Schmitzer, and F-X. Vialard.
\newblock {S}caling {A}lgorithms for {U}nbalanced {T}ransport {P}roblems.
\newblock \emph{arXiv:1607.05816 [math.OC]}, 2017.

\bibitem[Cohen et~al.(2021)Cohen, Luise, Terenin, Amos, and
  Deisenroth]{cohen2021aligning}
Samuel Cohen, Giulia Luise, Alexander Terenin, Brandon Amos, and Marc
  Deisenroth.
\newblock Aligning time series on incomparable spaces.
\newblock In \emph{International Conference on Artificial Intelligence and
  Statistics}, pages 1036--1044. PMLR, 2021.

\bibitem[Cuturi(2013)]{cuturi13}
M.~Cuturi.
\newblock {Sinkhorn Distances: Lightspeed Computation of Optimal Transport}.
\newblock In \emph{Neural Information Processing Systems}, 2013.

\bibitem[Cuturi(2011)]{cuturi11}
Marco Cuturi.
\newblock Fast global alignment kernels.
\newblock In \emph{{Proceedings of the 28th International Conference on
  International Conference on Machine Learning}}, {ICML'11}, pages 929--936,
  USA, 2011. Omnipress.

\bibitem[Cuturi and Blondel(2017)]{cuturi17}
Marco Cuturi and Mathieu Blondel.
\newblock Soft-dtw: a differentiable loss function for time-series.
\newblock In \emph{International Conference on Machine Learning}, 2017.

\bibitem[Cuturi and Peyr{\'e}(2018)]{cuturi2018}
Marco Cuturi and Gabriel Peyr{\'e}.
\newblock Semidual regularized optimal transport.
\newblock \emph{SIAM Review}, 60\penalty0 (4):\penalty0 941--965, 2018.

\bibitem[Cuturi et~al.(2007)Cuturi, Vert, Birkenes, and
  Matsui]{cuturi2007kernel}
Marco Cuturi, Jean-Philippe Vert, Oystein Birkenes, and Tomoko Matsui.
\newblock A kernel for time series based on global alignments.
\newblock In \emph{2007 IEEE International Conference on Acoustics, Speech and
  Signal Processing-ICASSP'07}, volume~2, pages II--413. IEEE, 2007.

\bibitem[Damodaran et~al.(2018)Damodaran, Kellenberger, Flamary, Tuia, and
  Courty]{Damodaran2018DeepJDOTDJ}
Bharath~Bhushan Damodaran, Benjamin Kellenberger, R{\'e}mi Flamary, D.~Tuia,
  and N.~Courty.
\newblock Deepjdot: Deep joint distribution optimal transport for unsupervised
  domain adaptation.
\newblock \emph{ArXiv}, abs/1803.10081, 2018.

\bibitem[Feydy et~al.(2017)Feydy, Charlier, Vialard, and Peyr{\'e}]{feydy17}
Jean Feydy, Benjamin Charlier, Fran{\c{c}}ois-Xavier Vialard, and Gabriel
  Peyr{\'e}.
\newblock Optimal transport for diffeomorphic registration.
\newblock pages 291--299, 2017.

\bibitem[Feydy et~al.(2018)Feydy, Séjourné, Vialard, Amari, Trouvé, and
  Peyré]{feydy19}
Jean Feydy, Thibault Séjourné, François-Xavier Vialard, Shun-ichi Amari,
  Alain Trouvé, and Gabriel Peyré.
\newblock Interpolating between optimal transport and mmd using sinkhorn
  divergences.
\newblock In \emph{Proceedings of the Twenty-Second International Conference on
  Artificial Intelligence and Statistics}, 10 2018.

\bibitem[Gramfort et~al.(2011)Gramfort, Papadopoulo, Baillet, and
  Clerc]{gramfort-etal:2011}
Alexandre Gramfort, Theodore Papadopoulo, Sylvain Baillet, and Maureen Clerc.
\newblock Tracking cortical activity from {M/EEG} using graph cuts with
  spatiotemporal constraints.
\newblock \emph{NeuroImage}, 54\penalty0 (3):\penalty0 1930 -- 1941, 2011.
\newblock ISSN 1053-8119.

\bibitem[Gramfort et~al.(2013)Gramfort, Luessi, Larson, Engemann, Strohmeier,
  Brodbeck, Goj, Jas, Brooks, Parkkonen, and Hämäläinen]{mne-python}
Alexandre Gramfort, Martin Luessi, Eric Larson, Denis Engemann, Daniel
  Strohmeier, Christian Brodbeck, Roman Goj, Mainak Jas, Teon Brooks, Lauri
  Parkkonen, and Matti Hämäläinen.
\newblock {MEG and EEG data analysis with MNE-Python}.
\newblock \emph{Frontiers in Neuroscience}, 7:\penalty0 267, 2013.
\newblock ISSN 1662-453X.
\newblock \doi{10.3389/fnins.2013.00267}.
\newblock URL
  \url{https://www.frontiersin.org/article/10.3389/fnins.2013.00267}.

\bibitem[Janati et~al.(2019)Janati, Cuturi, and Gramfort]{janati19}
Hicham Janati, Marco Cuturi, and Alexandre Gramfort.
\newblock Wasserstein regularization for sparse multi-task regression.
\newblock In \emph{Proceedings of the Twenty-First International Conference on
  Artificial Intelligence and Statistics}, volume~89 of \emph{Proceedings of
  Machine Learning Research}. PMLR, 16--19 Apr 2019.

\bibitem[Janati et~al.(2020{\natexlab{a}})Janati, Cuturi, and
  Gramfort]{janati20-aistats}
Hicham Janati, Marco Cuturi, and Alexandre Gramfort.
\newblock Spatio-temporal alignments: Optimal transport through space and time.
\newblock In \emph{Proceedings of the Twenty-third International Conference on
  Artificial Intelligence and Statistics}, Proceedings of Machine Learning
  Research. PMLR, 03--05 Jun 2020{\natexlab{a}}.

\bibitem[Janati et~al.(2020{\natexlab{b}})Janati, Cuturi, and
  Gramfort]{janati20-icml}
Hicham Janati, Marco Cuturi, and Alexandre Gramfort.
\newblock Debiased {S}inkhorn barycenters.
\newblock In Hal~Daumé III and Aarti Singh, editors, \emph{Proceedings of the
  37th International Conference on Machine Learning}, volume 119 of
  \emph{Proceedings of Machine Learning Research}, pages 4692--4701. PMLR,
  13--18 Jul 2020{\natexlab{b}}.
\newblock URL \url{http://proceedings.mlr.press/v119/janati20a.html}.

\bibitem[Kanwisher et~al.(1997)Kanwisher, McDermott, and Chun]{kanwisher97}
Nancy Kanwisher, Josh McDermott, and Marvin~M. Chun.
\newblock The fusiform face area: A module in human extrastriate cortex
  specialized for face perception.
\newblock \emph{Journal of Neuroscience}, 17\penalty0 (11):\penalty0
  4302--4311, 1997.
\newblock ISSN 0270-6474.
\newblock \doi{10.1523/JNEUROSCI.17-11-04302.1997}.
\newblock URL \url{https://www.jneurosci.org/content/17/11/4302}.

\bibitem[Liero et~al.(2016)Liero, Mielke, and Savar{\'e}]{liero16}
Matthias Liero, Alexander Mielke, and Giuseppe Savar{\'e}.
\newblock Optimal transport in competition with reaction: the
  {Hellinger}--{Kantorovich} distance and geodesic curves.
\newblock \emph{SIAM Journal on Mathematical Analysis}, 48\penalty0
  (4):\penalty0 2869--2911, 2016.

\bibitem[Saigo et~al.(2004)Saigo, Jean-Philippe, Vert, Ueda, and
  Akutsu]{saigo04}
Hiroto Saigo, Jean-Philippe, Vert, Nobuhisa Ueda, and Tatsuya Akutsu.
\newblock Protein homology detection using string alignment kernels.
\newblock \emph{Bioinformatics}, 20\penalty0 (11):\penalty0 1682–1689, 2004.

\bibitem[{Sakoe} and {Chiba}(1978)]{sakoe78}
H.~{Sakoe} and S.~{Chiba}.
\newblock Dynamic programming algorithm optimization for spoken word
  recognition.
\newblock \emph{IEEE Transactions on Acoustics, Speech, and Signal Processing},
  26\penalty0 (1):\penalty0 43--49, February 1978.
\newblock ISSN 0096-3518.

\bibitem[S{\'e}journ{\'e} et~al.(2019)S{\'e}journ{\'e}, Feydy, Vialard,
  Trouv{\'e}, and Peyr{\'e}]{sejourne19}
Thibault S{\'e}journ{\'e}, Jean Feydy, Fran{\c{c}}ois-Xavier Vialard, Alain
  Trouv{\'e}, and Gabriel Peyr{\'e}.
\newblock Sinkhorn divergences for unbalanced optimal transport.
\newblock \emph{arXiv preprint arXiv:1910.12958}, 2019.

\bibitem[Slotnick et~al.(1999)Slotnick, Klein, Carney, Sutter, and
  Dastmalchi]{slotnick99}
Scott~D. Slotnick, Stanley~A. Klein, Thom Carney, Erich Sutter, and Shahram
  Dastmalchi.
\newblock Using multi-stimulus vep source localization to obtain a retinotopic
  map of human primary visual cortex.
\newblock \emph{Clinical Neurophysiology}, 110\penalty0 (10):\penalty0 1793 --
  1800, 1999.
\newblock ISSN 1388-2457.
\newblock \doi{https://doi.org/10.1016/S1388-2457(99)00135-2}.
\newblock URL
  \url{http://www.sciencedirect.com/science/article/pii/S1388245799001352}.

\bibitem[Solomon et~al.(2015)Solomon, de~Goes, Peyr{\'e}, Cuturi, Butscher,
  Nguyen, Du, and Guibas]{solomon15}
J.~Solomon, F.~de~Goes, G.~Peyr{\'e}, M.~Cuturi, A.~Butscher, A.~Nguyen, T.~Du,
  and L.~Guibas.
\newblock Convolutional {Wasserstein} distances: Efficient optimal
  transportation on geometric domains.
\newblock \emph{ACM Trans. Graph.}, 34\penalty0 (4):\penalty0 66:1--66:11, July
  2015.
\newblock ISSN 0730-0301.

\bibitem[Stanley(2011)]{stanley11}
Richard~P. Stanley.
\newblock \emph{Enumerative Combinatorics: Volume 1}.
\newblock Cambridge University Press, New York, NY, USA, 2nd edition, 2011.
\newblock ISBN 1107602629, 9781107602625.

\bibitem[Thorpe et~al.(2017)Thorpe, Park, Kolouri, Rohde, and Slep{\v
  c}ev]{thorpe17}
Matthew Thorpe, Serim Park, Soheil Kolouri, Gustavo~K Rohde, and Dejan Slep{\v
  c}ev.
\newblock A transportation l(p) distance for signal analysis.
\newblock \emph{Journal of mathematical imaging and vision}, 59\penalty0
  (2):\penalty0 187--210, 10 2017.

\bibitem[Vayer et~al.(2020)Vayer, Chapel, Courty, Flamary, Soullard, and
  Tavenard]{vayer2020}
Titouan Vayer, Laetitia Chapel, Nicolas Courty, R{\'e}mi Flamary, Yann
  Soullard, and Romain Tavenard.
\newblock Time series alignment with global invariances.
\newblock \emph{arXiv preprint arXiv:2002.03848}, 2020.

\bibitem[Wakeman and Henson(2015)]{ds117}
D.G. Wakeman and R.N.A. Henson.
\newblock A multi-subject, multi-modal human neuroimaging dataset.
\newblock \emph{Scientific Data}, 2\penalty0 (150001), 2015.

\bibitem[Williams et~al.(2006)Williams, M.Toussaint, and Storkey.]{williams}
B.H. Williams, M.Toussaint, and A.J. Storkey.
\newblock Extracting motion primitives from natural handwriting data.
\newblock In \emph{ICANN}, volume~2, page 634–643, 2006.

\bibitem[Yarkoni(2014)]{neurosynth}
Tal Yarkoni.
\newblock Neurosynth core tools v0.3.1, May 2014.
\newblock URL \url{https://doi.org/10.5281/zenodo.9925}.

\end{thebibliography}


\cleardoublepage

\end{document}